\documentclass[sigconf, 10pt]{acmart}

\usepackage{booktabs} % For formal tables

\usepackage{subfigure}
\usepackage{bm}
\usepackage{algorithm, algorithmic}
\usepackage{array,enumerate}

\makeatletter
\newcommand{\thickhline}{%
    \noalign {\ifnum 0=`}\fi \hrule height 1pt
    \futurelet \reserved@a \@xhline
}
\newcolumntype{"}{@{\hskip\tabcolsep\vrule width 1pt\hskip\tabcolsep}}
\makeatother

\usepackage{epstopdf}
\setcopyright{rightsretained}
\acmDOI{10.475/123_4}

\acmISBN{123-4567-24-567/08/06}

\acmConference[]{}{}{} 
\acmYear{2018}
\copyrightyear{2018}

\acmArticle{4}
\acmPrice{15.00}
\begin{document}
\renewcommand{\arraystretch}{1.3}
\title{Time Series Segmentation through Automatic Feature Learning}

\author{Wei-Han Lee}
\affiliation{%
  \institution{Princeton University}
}
\email{weihanl@princeton.edu}
\author{Jorge Ortiz}
\affiliation{%
  \institution{IBM Research}
}
\email{jjortiz@us.ibm.com}
\author{Bongjun Ko}
\affiliation{%
  \institution{IBM Research}
  }
\email{bongjun_ko@us.ibm.com}

\author{Ruby Lee}
\affiliation{%
  \institution{Princeton University}
}
\email{rblee@princeton.edu}
%\affiliation{%
% \institution{Rajiv Gandhi University}
%\affiliation{%
%  \institution{Tsinghua University}
%  \streetaddress{30 Shuangqing Rd}
%  \country{China}
%\affiliation{%
%  \institution{Palmer Research Laboratories}
%  \streetaddress{8600 Datapoint Drive}
%\affiliation{\institution{The Th{\o}rv{\"a}ld Group}}
%\email{jsmith@affiliation.org}
%
%\author{Julius P.~Kumquat}
%\affiliation{\institution{The Kumquat Consortium}}
%\email{jpkumquat@consortium.net}
%
%% The default list of authors is too long for headers.

\begin{abstract}
Internet of things (IoT) applications have become increasingly popular in recent years, with applications ranging from building energy monitoring to personal health tracking and activity recognition. In order to leverage these data, automatic knowledge extraction -- whereby we map from observations to interpretable states and transitions -- must be done at scale. As such, we have seen many recent IoT data sets include annotations with a human expert specifying states, recorded as a set of boundaries and associated labels in a data sequence. These data can be used to build automatic labeling algorithms that produce labels as an expert would. Here, we refer to human-specified boundaries as \emph{breakpoints}. Traditional changepoint detection methods only look for statistically-detectable boundaries that are defined as abrupt variations in the generative parameters of a data sequence. However, we observe that \emph{breakpoints} occur on more subtle boundaries that are non-trivial to detect with these statistical methods. In this work, we propose a new unsupervised approach, based on deep learning, that outperforms existing techniques and learns the more subtle, \emph{breakpoint} boundaries with a high accuracy. Through extensive experiments on various real-world data sets -- including human-activity sensing data, speech signals, and electroencephalogram (EEG) activity traces -- we demonstrate the effectiveness of our algorithm for practical applications. Furthermore, we show that our approach achieves significantly better performance than previous methods.
\end{abstract}

\begin{CCSXML}
<ccs2012>
<concept>
<concept_id>10002951.10003227</concept_id>
<concept_desc>Information systems~Information systems applications</concept_desc>
<concept_significance>500</concept_significance>
</concept>
<concept>
<concept_id>10010147.10010257.10010293</concept_id>
<concept_desc>Computing methodologies~Machine learning approaches</concept_desc>
<concept_significance>500</concept_significance>
</concept>
<concept>
<concept_id>10010147.10010257.10010321</concept_id>
<concept_desc>Computing methodologies~Machine learning algorithms</concept_desc>
<concept_significance>500</concept_significance>
</concept>
</ccs2012>
\end{CCSXML}

\ccsdesc[500]{Information systems~Information systems applications}
\ccsdesc[500]{Computing methodologies~Machine learning approaches}
\ccsdesc[500]{Computing methodologies~Machine learning algorithms}

\keywords{Time-series Segmentation; Deep Learning; Automatic Feature Extraction;}

\maketitle

\begin{figure*}[!t]
\centering
\subfigure[]{
\label{fig:visual_compare_Gamma}
\includegraphics[width=3.7in,height=1.5in]{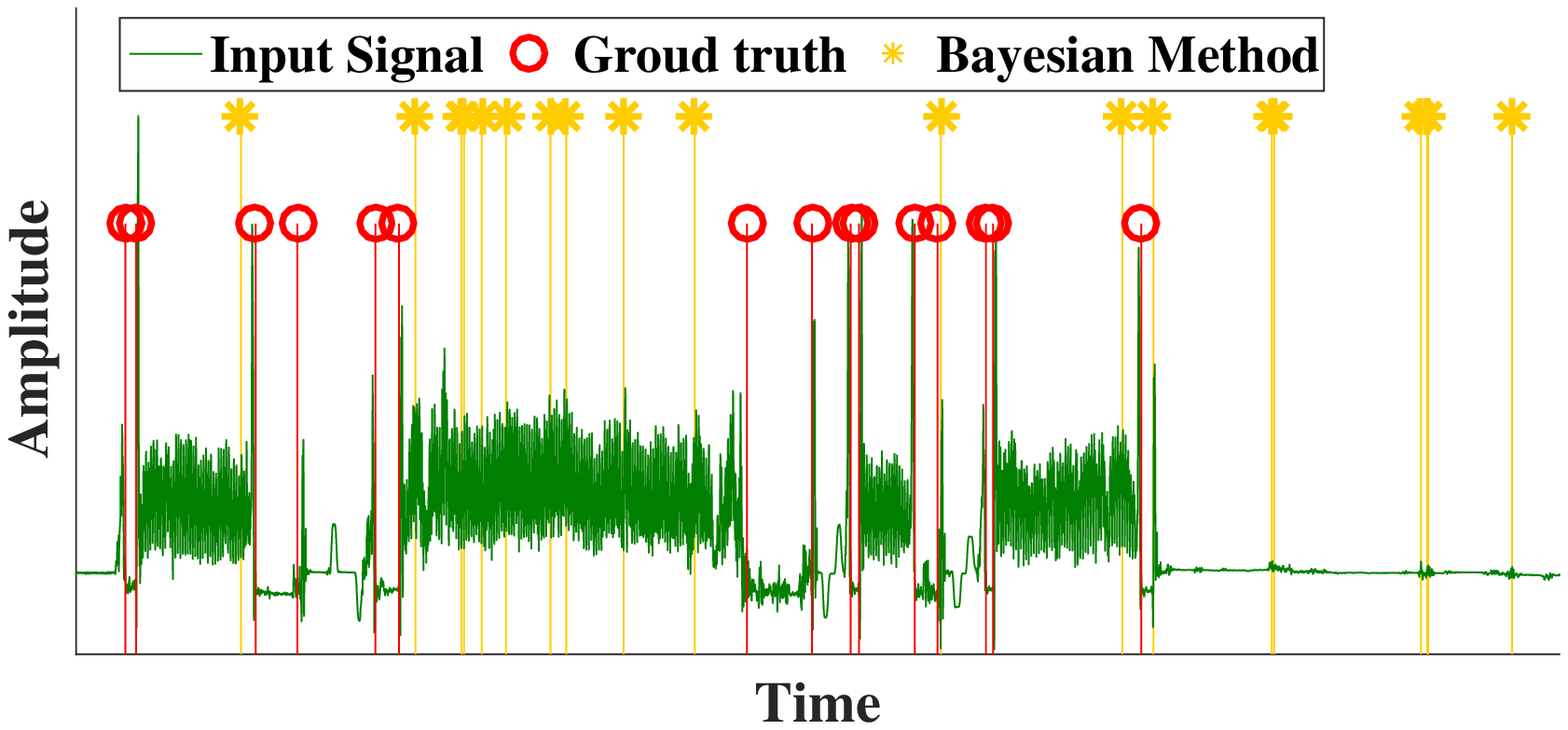}}
\hspace{-5em}
\subfigure[]{
\label{fig:visual_compare_Gaussian}
\includegraphics[width=3.7in,height=1.5in]{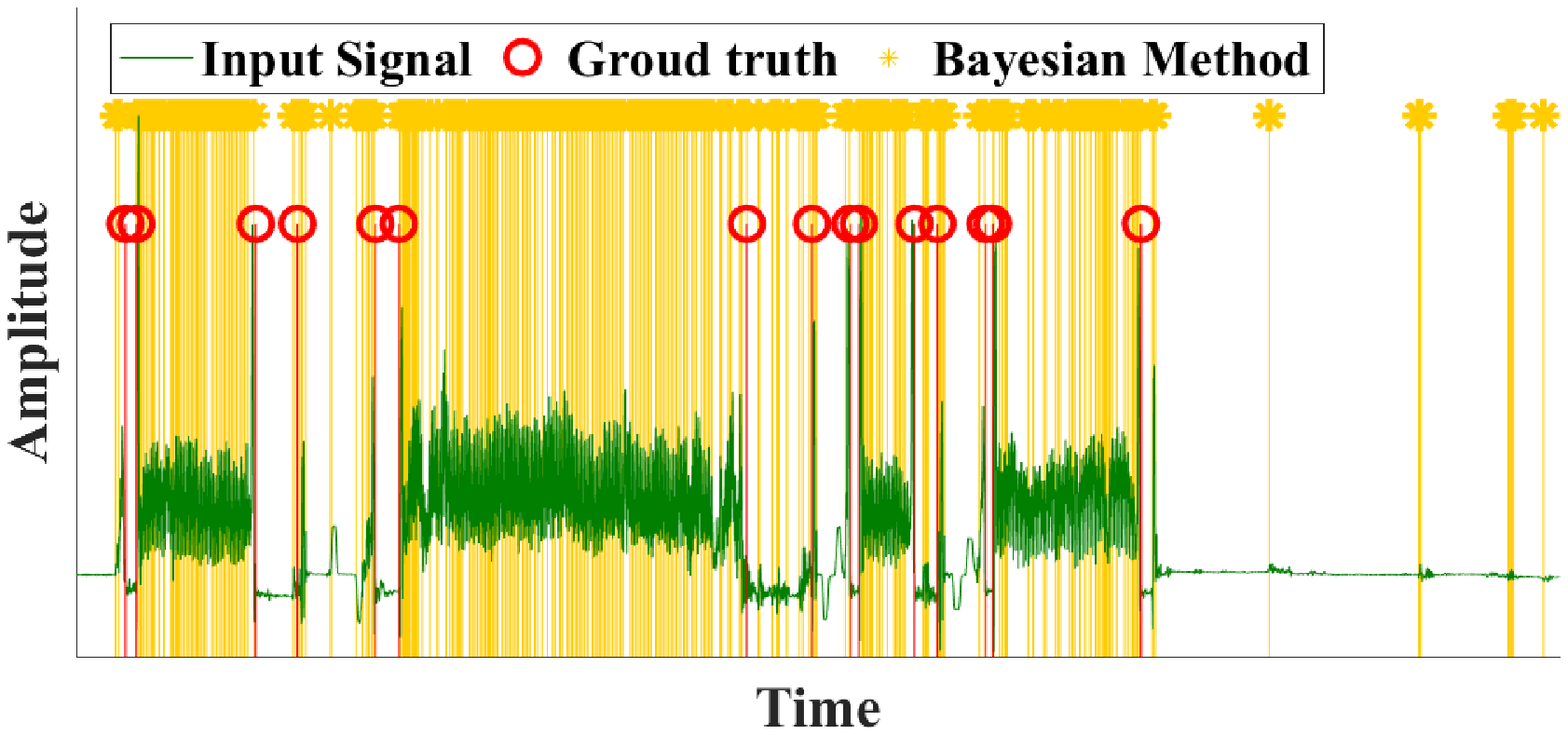}}
\subfigure[]{
\label{fig:visual_compare_Our}
\includegraphics[width=3.7in,height=1.5in]{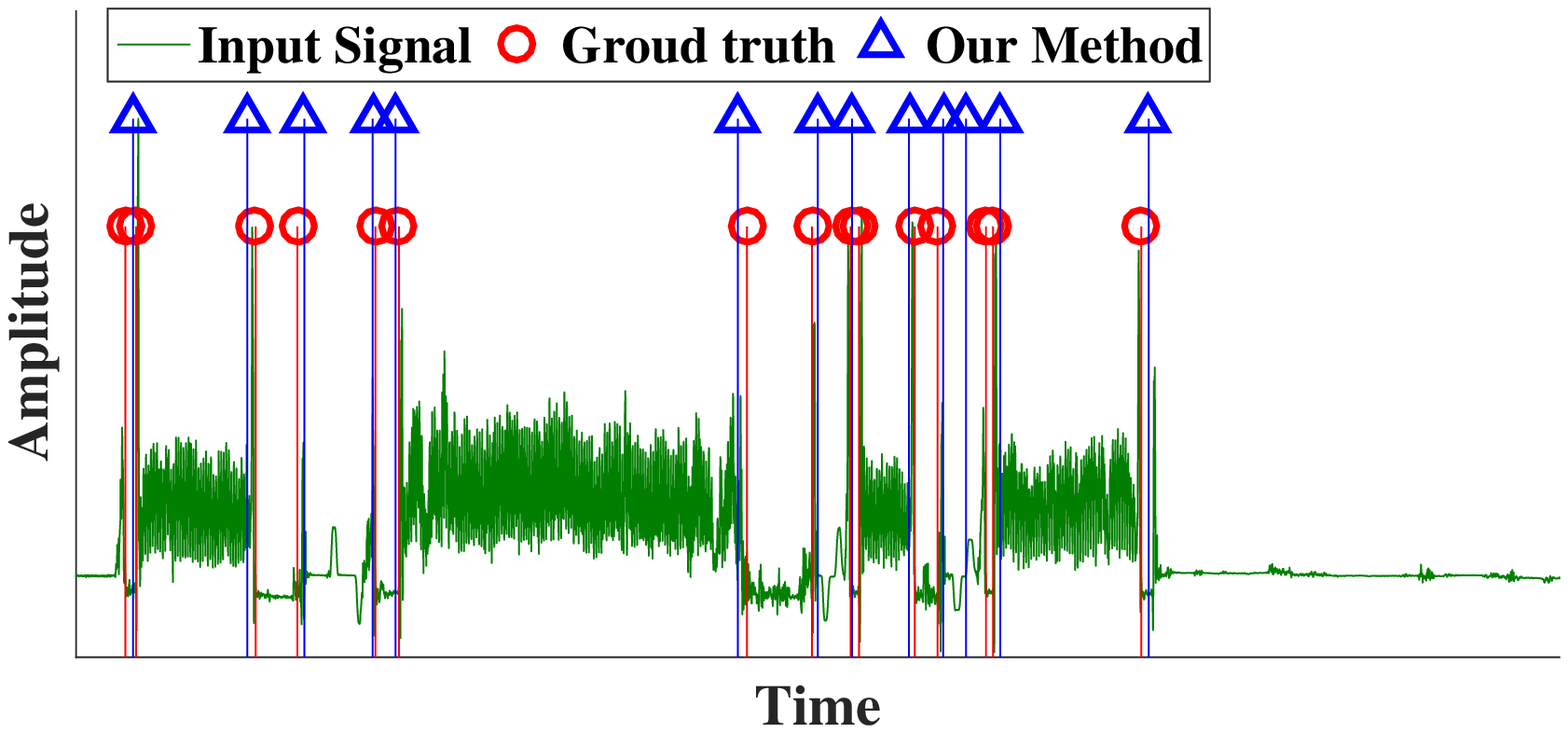}}
\caption{The performance of breakpoint detection under different methods using a smartphone sensor data set for activity recognition~\cite{crowdsignals}. The green line represents the original signal and the red circle line is the ground truth of breakpoints. The yellow star lines in (a) and (b) represent the detected breakpoints by using existing Bayesian method with prior distribution of Gamma and Gaussian, respectively~\cite{adams2007bayesian}. The blue triangle lines in (c) represent the detected breakpoints by using our method. We can see that our method significantly outperforms the previous approaches in finding breakpoints for real-world applications.}
\label{fig:visual_compare}
\end{figure*}

\section{Introduction}
Changepoint detection is an important, fundamental technique used in the analysis of time series data. It has been generally applied in analyzing stock data~\cite{hsu1982bayesian,oh2002analyzing,hasan2014information}, sensor data from Internet of things (IoT) deployments~\cite{ramos2016anomalies,xie2013sequential}, physiological data analysis~\cite{chen2011parametric,rosenfield2010change}, and many others~\cite{fryzlewicz2014wild,jaruskova1997some,siris2004application,wang2004change}. Changepoint detection is fundamental for discovering how distinct sequences of values might be associated with states in a process that are not directly observable. By examining changepoints, analysts can build models of those sequences or look for patterns of sequences across multiple data sets. Changepoint detection is a fundamental primitive for building state-space process models.

As the amount of available data grows, we observe that a large fraction of it is now annotated with human labels provided by a domain expert. These labels are useful for modeling latent states and state-transition sequences. By examining the temporal boundaries for states as specified in the annotated data, analysts can look for similar transition patterns and feed more complex models that capture the relationship between those states. For example, many IoT mobile phone applications infer a user's activity using onboard sensors. In order to train  these models, the user must provide information about their activity. This is recorded as an annotation in the data with a start time and end time. Similarly, experts spend a great deal of time annotating electrocardiogram (ECG) data with labels that separate traces into  the various cardiac states of the patient.

Changepoints are abrupt changes in the trends of a data sequence. Bayesian techniques~\cite{ray2002bayesian,adams2007bayesian,barry1993bayesian,bai1997estimation,erdman2008fast} discover these by looking for changes in the parameters of the distribution that generates the sequence. Considering the generality of this problem, many techniques exist in the literature.
These techniques attempt to capture the generative process through a pre-determined model and aim to look for changes in the parameters of the generative process.
For learning expert-specified boundaries, these models often fail -- since such changes are not easily captured by a pre-specified model of the generative process and changepoints do not typically fall along parameter-shift boundaries.

The changepoints specified by experts often arise when the state transition is a function of latent temporal properties of an underlying process that are difficult to capture in a pre-specified model. These rules are encoded as latent features in these traces and practically impossible to detect with the existing generative-model based changepoint detection algorithms~\cite{ray2002bayesian,adams2007bayesian,barry1993bayesian,bai1997estimation,erdman2008fast}. We observe that existing methods do a poor job of identifying human-specified changepoints. In summary, existing changepoint detection methods have two main weaknesses: 1) they rely on a prior parametric model of the time series data, and 2) they often utilize simple features extracted from the input data such as the mean, variance, spectrum, etc. 
Therefore, previous methods can only discover statistically-detectable boundaries. To differentiate from these statistically-detectable changepoints, we hereafter refer to the human-specified changepoints as \emph{breakpoints}. Furthermore, we propose a novel algorithm that uses deep learning techniques to detect \emph{breakpoints} without any prior assumptions about the generative process. Our method automatically learns the features that are most useful to represent the input data and thus can discover hidden structure in real-world time series data. Note that our approach has broad applicability for general changepoint detection even outside of its application to breakpoint detection as considered in this paper.

Figure~\ref{fig:visual_compare} shows a comparison of our approach to a Bayesian changepoint detection technique from the literature~\cite{adams2007bayesian} by using a smartphone sensor data set for activity detection~\cite{crowdsignals}. Note that even after careful tuning of the  parameters in~\cite{adams2007bayesian}, their method still could not accurately detect these breakpoint boundaries. Furthermore, their technique is sensitive to parameter changes and one can easily over or under estimate the number of breakpoints. Moreover, it is not at all clear how to adapt the parameters to capture the statistical properties of a true segment.  In comparison, our approach learns this automatically.  We will explain how we choose the hyperparameters of our model through a simple set of heuristics we learn through direct observation and analysis on real-world traces.

In summary, we make the following contributions:
\begin{itemize}
\item We introduce a new kind of changepoint called a \emph{breakpoint} and show that it is nearly impossible to detect with existing changepoint detection techniques.
\item We propose a novel method that utilizes deep learning to automatically learn useful features that represent data sequences generated from an expert-specified sequential segments. Our technique does not rely on the assumption that the changepoints are caused by abrupt changes of the parameters in the generative process as previous methods do, making it applicable for a broad coverage of real-world applications. 
\item We demonstrate the effectiveness of our method through extensive experimental analysis using multiple real-world data sets. Furthermore, we show how to choose the hyperparameters of our model from a simple heuristic derived from the association between statistical properties of the data and the performance of our model. These experimental analysis demonstrate that our approach can serve as a key enabler for real-world applications. 
\item Furthermore, we compare our method with several existing approaches and introduce a new metric that measures the effectiveness of a changepoint detection scheme with respect to accuracy in the number of predicted changepoints and their overlap with true changepoint coordinates. The experimental results show the significant advantage our approach has over existing methods. 
\end{itemize}

\section{Related Work}
Changepoint detection has attracted researchers in the statistics and data mining communities for decades \cite{lee2017implicit,hsu1982bayesian,oh2002analyzing,hasan2014information,ramos2016anomalies,xie2013sequential,chen2011parametric,rosenfield2010change,reeves2007review,wang2011non,yamanishi2004line}. Changepoint detection techniques have been applied in various applications such as stock data analysis~\cite{hsu1982bayesian,oh2002analyzing,hasan2014information}, sensor data analysis in IoT systems~\cite{ramos2016anomalies,xie2013sequential,lee2017implicit}, physiological data analysis~\cite{chen2011parametric,rosenfield2010change}, climate change detection \cite{reeves2007review}, genetic time-series analysis \cite{wang2011non}, and intrusion detection in computer networks \cite{yamanishi2004line}. 

One important thread of changepoint detection compares the probability distributions of time-series samples over the past and present intervals. As both the intervals move forward, a typical strategy is to issue an alarm for a changepoint when the two distributions become `significantly' different. Various changepoint detection methods apply this strategy such as the cumulative sum method~\cite{basseville1993detection}, the generalized likelihood-ratio method \cite{gustafsson1996marginalized} and the change finder method \cite{yamanishi2002unifying}. A similar strategy has also been employed in novelty detection \cite{guralnik1999event} and outlier detection \cite{hido2011statistical}.

Another common thread is the subspace method \cite{jurdjevic2007change,kawahara2007change,moskvina2001application}. By using a pre-designed time-series model, a subspace is discovered by principal component analysis (PCA) from trajectories in the past and present intervals, and their dissimilarity is measured by the distance between the subspaces. However, a key challenge for these methods is how to accurately estimate the density. To solve this problem, previous work tries to estimate the density-ratio instead of the density itself. The rationale is that knowing the two densities implies knowing the density ratio, but not vice versa since such a decomposition is not unique. Thus, direct density-ratio estimation is substantially easier than density estimation \cite{sugiyama2012density}. Following this idea, methods of direct density-ratio estimation have been developed such as the kernel mean matching method \cite{gretton2009covariate}, the logistic-regression method \cite{bickel2007discriminative} and the Kullback-Leibler importance estimation procedure (KLIEP)~\cite{sugiyama2008direct}.

Most of the existing changepoint detection methods are fundamentally limited in the type of changepoints they can detect because: 1) they rely on pre-designed parametric models such as underlying probability distributions \cite{basseville1993detection,gustafsson2000adaptive}, auto-regressive models \cite{yamanishi2002unifying}, and state-space models \cite{jurdjevic2007change,kawahara2007change}. 
In practice, we observe that the pre-specified parametric models are difficult to parameterize so that the predicted changepoints align with annotation boundaries in the data; and 2) they utilize simple statistical properties such as the mean, variance, and spectrum~\cite{basseville1993detection,gustafsson2000adaptive,sugiyama2012density,moskvina2001application,brodsky2013nonparametric} to serve as features for changepoint detection. However, these features do not generalize well and performance varies substantially across different data sets. Therefore, existing changepoint detection methods can only detect statistically-detectable boundaries (and not the human-specified breakpoints). 

To overcome these problems, we propose a novel breakpoint detection scheme to detect human-specified boundaries, through learning the most representative features specific to the input time-series data and exploiting these features for segmentation.
Our technique utilizes an autoencoder model~\cite{hinton2006reducing,lee2008sparse,vincent2008extracting} to automatically learn the features that are most useful to represent the input time-series data, which can be generally applied to a broad coverage of real-world applications.

\begin{figure*}
\centering
\includegraphics[width=7in,height=2in]{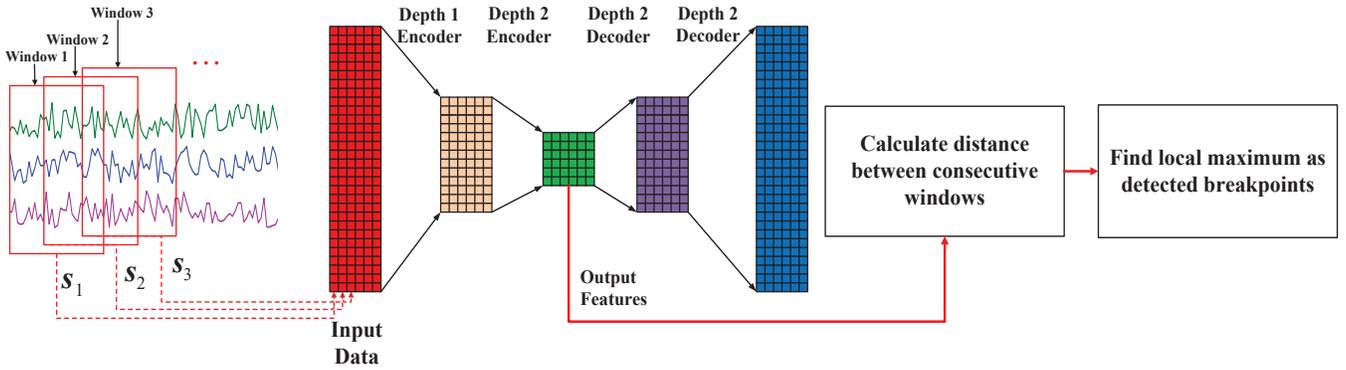}
\caption{Pipeline for our breakpoint detection system. We first segment the input data into a series of windows and then apply autoencoder models in deep learning to extract representative features for the input data. These extracted features can then be utilized to calculate the distance between consecutive windows and the timestamps corresponding to local-maximal distance can be detected as breakpoints.}
\label{fig:pipeline}
\end{figure*}
\section{Autoencoder-based \\ Breakpoint Detection}\label{our_mechanism}
In this section, we describe our breakpoint detection approach, which utilizes a deep autoencoder model to extract the most representative features in time-series data. The key idea is that the autoencoder models in deep learning techniques can automatically and effectively extract the unique features specific to the input data without making any prior assumption about the generative process which produced the data. In this way, we can obtain a deeper understanding about the temporal dynamics of the input data and achieve better performance for detecting user-specified breakpoints. The end-to-end pipeline of our method is shown in Figure~\ref{fig:pipeline}, and detailed steps are described below.

\subsection{Data Preprocessing}
For a given time series, consisting of $N_c$ channels (such as different sensors in an IoT system) across $T$ timestamps, the input data matrix $\bm{IDM}\in\mathbb{R}^{N_c\times T}$ is a real matrix where $\bm{IDM}(i,j)$ is the measurement recorded by the $i$-th channel at the $j$-th timestamp. To fully explore the temporal characteristics of the data, we follow common practice and partition it into a series of segments according to a user-specified time window size, $N_w$.
For the $t$-th ($t=1,2,\cdots,T/N_w$) window, we stack all the recordings within it to form a column vector which is denoted by ${\bm{s}}_t \in \mathbb{R}^{N_c N_{w}\times 1}$. 
The input data matrix can thus be reformulated as ${\bm{S}}=[{\bm{s}}_1; {\bm{s}}_2; \cdots]$. 
Note that the segmented data may have some overlapping recordings as shown in Figure~\ref{fig:pipeline}.

\subsection{Automatic Feature Extraction}\label{autoencoder}
Deep learning models learn multi-layer transformations from the input data to the output representations, which is more powerful in feature extraction than hand-crafted shallow models. Moreover, deep learning models can progressively capture more compact features at higher layers, corresponding to the hierarchical human vision systems. Among the building blocks of these models, autoencoders \cite{hinton2006reducing,lee2008sparse,vincent2008extracting} automatically learn a non-parametric feature mapping function by minimizing the reconstruction error between the input and its reconstructed output. Therefore, we first aim to explore the autoencoder techniques to extract features that are most useful for representing the input data.

Autoencoders \cite{hinton2006reducing,lee2008sparse,vincent2008extracting}, as the name suggests, consist of two stages: encoding and decoding. A single-layer autoencoder, which is a kind of neural network consisting of only one hidden layer, aims to find a common feature basis from the input data. It was first used to reduce dimensionality by setting the number of the extracted features less than the input. If the dimension of the encoding output is set higher, the encoding result will be enriched and more expressive. The autoencoder model is usually trained by the \emph{back-propagation} techniques~\cite{bottou1991stochastic} in an unsupervised manner, aiming at minimizing the error of the reconstructed results from the original inputs. By stacking multiple autoencoders, the deep autoencoders will generate more compact and higher-level semantic features which are beneficial for feature representation.

In autoencoder model, the encoder is a function $E_{nc}(\cdot)$ that maps the input data ${\bm{s}}\in\mathbb{R}^{d_{{\bm{s}}}\times 1}$ to ${\bm{f}}\in\mathbb{R}^{d_{\bm{f}}\times 1}$ hidden units to obtain the feature representation as
\begin{equation}\label{encoding}
{\bm{f}}=E_{nc}({\bm{s}})=g_{enc}({\bm{W}}{\bm{s}}+{\bm{b}}_e)
\end{equation} 

where $g(\cdot)$ is a nonlinear activation function, typically a sigmoid function:
\begin{equation}
g(r)=\frac{1}{1+e^{-r}}
\end{equation} 
or a hyperbolic tangent function:
\begin{equation}
 g(r)=tanh(r)=\frac{e^{r}-e^{-r}}{e^{r}+e^{-r}}
\end{equation}
 
The parameters of the encoder consist of a weight matrix $\bm{W}\in\mathbb{R}^{d_{\bm{f}}\times d_{\bm{s}}}$ and a bias vector ${\bm{b}}_e\in\mathbb{R}^{{{d}}_{{\bm{f}}}\times 1}$.

The decoder function $D_{ec}(\cdot)$ maps the outputs of the hidden units (feature representations) back to the original input space according to
\begin{equation}
\widetilde{\bm{s}}=D_{ec}({\bm{f}})=g_{dec}({\bm{W}}^\prime{\bm{f}}+{\bm{b}}_d)
\end{equation} 
\indent where $g_{dec}(\cdot)$ is usually the same form as that in the encoder. The parameters of the decoder also consist of a  weight matrix ${\bm{W}}^\prime\in\mathbb{R}^{d_{{\bm{s}}}\times d_{\bm{f}}}$ and a bias vector ${\bm{b}}_d\in\mathbb{R}^{d_{{\bm{s}}}\times 1}$. 

Here, we choose both the encoding and decoding activation function to be \emph{sigmoid} function and only consider the tied weights case, in which ${\bm{W}}^\prime={\bm{W}}^T$ (where ${\bm{W}}^T$ is the transpose of ${\bm{W}}$) as most existing deep learning methods do~\cite{hinton2006reducing,lee2008sparse,vincent2008extracting}.

The objective of autoencoder model is to minimize the error of the reconstructed result $D_{ec}(E_{nc}({\bm{s}}_t))$ from the input data ${\bm{s}}_t$ as 
\begin{equation}\label{cost}
\begin{aligned}
\min J({\bm{W}},{\bm{b}}_e,{\bm{b}}_d,{\bm{W}}^\prime) &= \min\sum_{t=1}^{T/N_w} L({\bm{s}}_t,D_{ec}(E_{nc}({\bm{s}}_t))) \\
												  &+ \lambda\sum_{t,i}{\bm{W}}_{t,i}^2
\end{aligned}
\end{equation}

where $L({\bm{u}},{\bm{v}})$ is a loss function which is usually decided according to the input range. Typical error functions include the cross-entropy loss:
\begin{equation}
L({\bm{u}},{\bm{v}})=\sum_{i=1}^{d_u} u_i\log(v_i)+(1-u_i)\log(1-v_i)
\end{equation}  
or the square loss:
\begin{equation}
L({\bm{u}},{\bm{v}})=\|{\bm{u}-{\bm{v}}}\|^2
\end{equation}

The second term in Eq.~\ref{cost} is a regularization term (also called a weight decay term) that tends to decrease the magnitude of the weights, and helps prevent overfitting~\cite{hinton2006reducing,lee2008sparse,vincent2008extracting}.

\begin{figure*}
\centering
\subfigure[Synthetic data using Generative Models]{
\includegraphics[width=3in,height=2.5in]{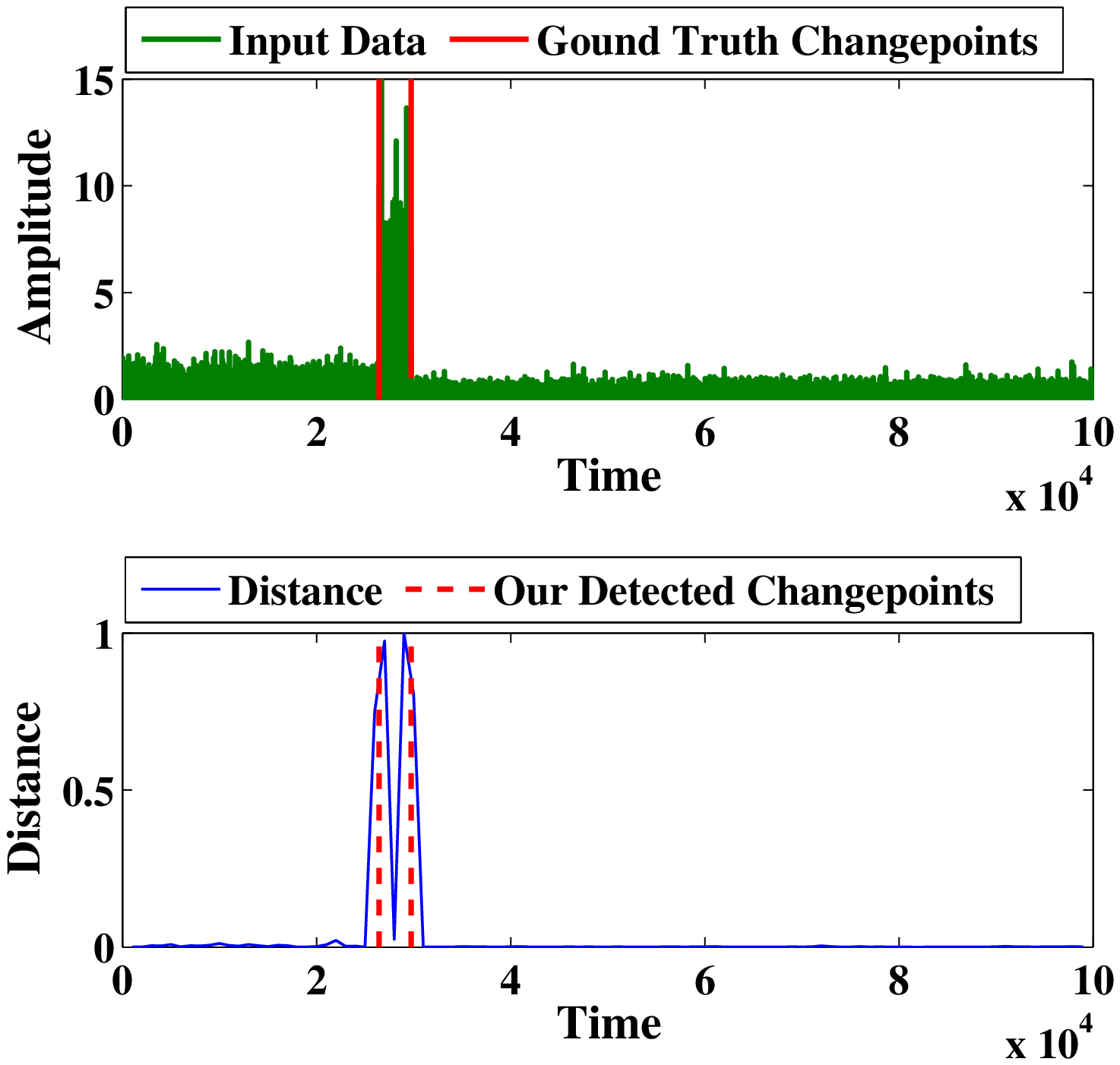}}
\subfigure[Crowdsignal.io Data Set]{
\includegraphics[width=3in,height=2.5in]{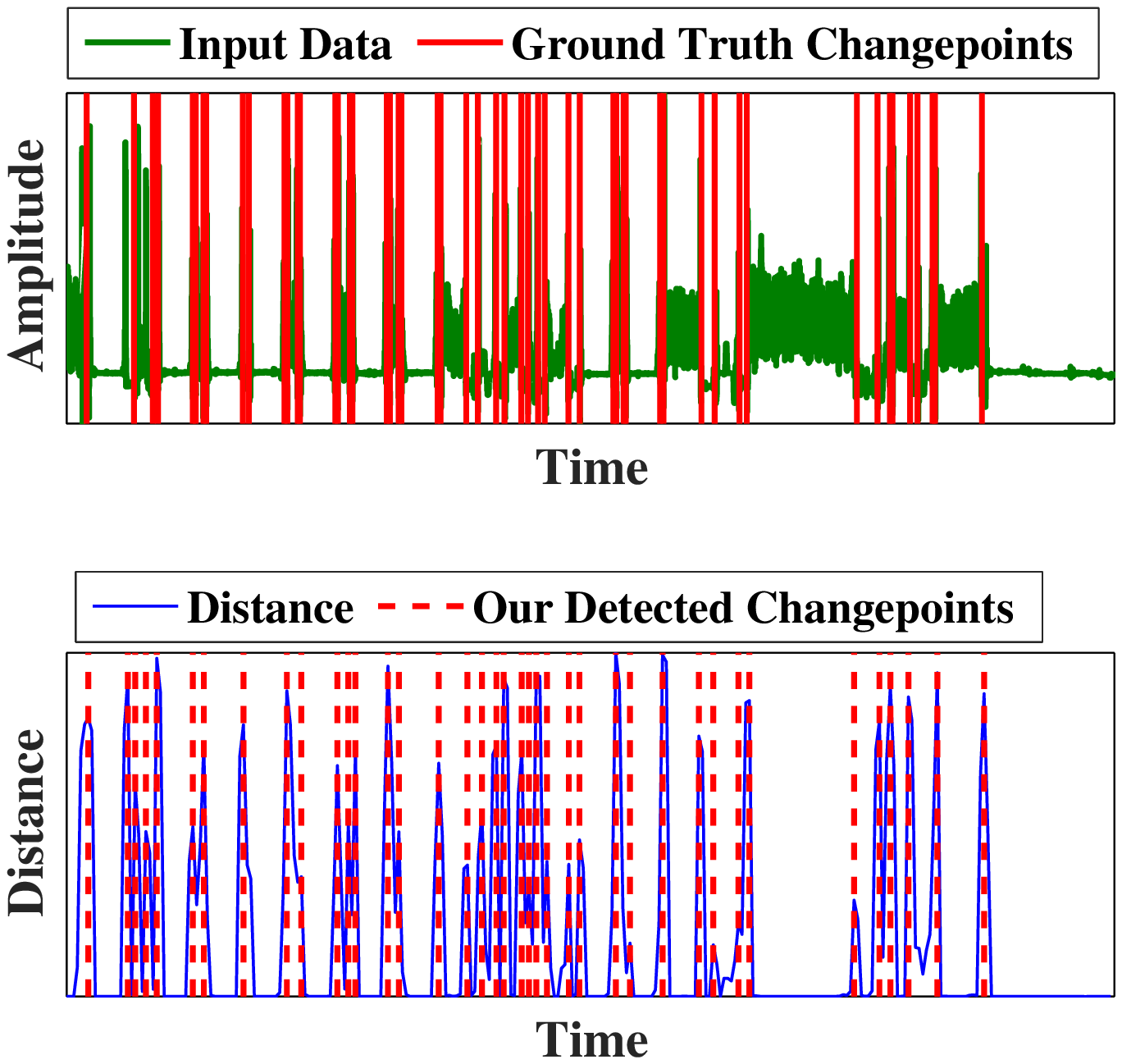}}
\caption{Experiments of our approach on a synthetic data produced using generative models and the real-world Crowdsignal.io  data set \cite{crowdsignals}. The upper figures show the raw input signals (shown as green lines) with the ground truth of changepoints shown as red lines. The bottom figures show the distance between features in two consecutive time windows (shown as blue lines) with the detected changepoints shown as red dash lines. We can see that our method is effective in detecting both statistically-detectable changepoints (using a synthetic data) and human-specified breakpoints (using the real-world Crowdsignal.io data).}
\label{fig:data3}
\end{figure*}

\noindent{\bf Model Learning and Stacking:} Our objective for feature representation is to minimize $J({\bm{W}},{\bm{b}}_e,{\bm{b}}_d,{\bm{W}}^\prime)$ in Eq.~\ref{cost} with respect to ${\bm{W}},{\bm{b}}_e,{\bm{b}}_d,{\bm{W}}^\prime$. We explore the stochastic gradient descent technique~\cite{bottou1991stochastic}  to solve such an optimization problem, which has been shown to perform fairly well in practice. To train our autoencoder network, we first initialize each parameter ${\bm{W}},{\bm{b}}_e,{\bm{b}}_d,{\bm{W}}^\prime$ to a small random value near zero, and then apply the {stochastic gradient descent} technique for iterative optimization. Note that we can compute the exact gradient for this objective function with respect to each variable ${\bm{W}},{\bm{b}}_e,{\bm{b}}_d,{\bm{W}}^\prime$.

\begin{algorithm}[!t]
\renewcommand{\algorithmicrequire}{\textbf{Input:}}
\renewcommand\algorithmicensure {\textbf{Output:} }
\begin{algorithmic}[1]
\REQUIRE~~{The input data $\{{\bm{s}}_t\}_{t=1}^{T/N_w}$, $\alpha$ is the learning rate;}\\
\ENSURE {The set of detected breakpoints $\mathbb{C}$};\\
       {
      % \vspace{1em}  
       /*****Optimize ${\bm{W}},{\bm{b}}_e, {\bm{b}}_d, {\bm{W}}^\prime$ for Feature Extraction*****/ \\ 
       1. Initialize ${\bm{W}},{\bm{b}}_e, {\bm{b}}_d$ randomly and set the tied weights \\
       \quad${\bm{W}}^\prime={\bm{W}}^T$ according to~\cite{hinton2006reducing,lee2008sparse,vincent2008extracting}; \\
       \quad Initialize the set of breakpoints as $\mathbb{C}=\emptyset$;\\
       2. {\bf{For each iteration}} $=1,2,3\cdots$ {\bf{do}}\\
       3.\quad\quad Set ${{\Delta}} {\bm{W}}=0, {{\Delta}} {\bm{b}}_e=0, {\bm{\Delta}} {\bm{b}}_d=0, {\bm{\Delta}}{\bm{W}}^\prime=0$;\\
       4.\quad\quad Compute the partial derivatives with respect to \\
       \quad\quad \hspace{0.2cm} ${\bm{W}},{\bm{b}}_e,{\bm{b}}_d, {\bm{W}}^\prime$ as
       \begin{equation}
       \begin{aligned}
       &~~~\Delta{\bm{W}}=\frac{\partial J({\bm{W}},{\bm{b}}_e,{\bm{b}}_d,{\bm{W}}^\prime)}{\partial {\bm{W}}}\\
       &~~~\Delta{\bm{b}}_e=\frac{\partial J({\bm{W}},{\bm{b}}_e,{\bm{b}}_d,{\bm{W}}^\prime)}{\partial {\bm{b}}_e}\\
       &~~~\Delta{\bm{b}}_d=\frac{\partial J({\bm{W}},{\bm{b}}_e,{\bm{b}}_d,{\bm{W}}^\prime)}{\partial {\bm{b}}_d}
       \\
       &~~~\Delta{\bm{W}}^\prime=\frac{\partial J({\bm{W}},{\bm{b}}_e,{\bm{b}}_d,{\bm{W}}^\prime)}{\partial {\bm{W}}^\prime}
       \end{aligned}
       \end{equation} 
       5.\quad\quad Update ${\bm{W}},{\bm{b}}_e,{\bm{b}}_d, {\bm{W}}^\prime$ by gradient descent as
       \begin{equation}
       \begin{aligned}
       &{\bm{W}}:={\bm{W}}-\alpha\Delta{\bm{W}}\\
       &{\bm{b}}_e:={\bm{b}}_e-\alpha\Delta{\bm{b}}_e\\
       &{\bm{b}}_d:={\bm{b}}_d-\alpha\Delta{\bm{b}}_d\\
       &{\bm{W}}^\prime:={\bm{W}}^\prime-\alpha\Delta{\bm{W}}^\prime
       \end{aligned}
       \end{equation}
       
      %   \vspace{1em}
      /****************** Breakpoint Detection *****************/\\  
       6. {\bf For each segmented time window} ${\bm{s}}_t$ {\bf do}\\
       7. \quad\quad Extract features ${\bm{f}}_t$ according to Eq.~\ref{encoding};\\
       8. \quad\quad Compute distances ${Dist}_{t}$ between consecutive\\ 
       \quad\quad\quad feature vectors according to Eq.~\ref{distance};\\
       9. \quad\quad{\bf If ${Dist}_t$ is a \emph{local-maximal distance} do}\\
       10.\quad\quad\quad Classify $t$ as a breakpoint, i.e., $\mathbb{C}\gets \mathbb{C}\cup t$;\\
       }
\caption{Our Deep Learning Based Breakpoint Detection Approach.}
\label{alg1}
\end{algorithmic}
\end{algorithm}

\indent Although the stochastic gradient descent method is effective for solving Eq. \ref{cost}, the learnt result heavily relies on the seeds used to initialize the optimization process. Therefore, we use multiple hidden layers to stack the model in order to achieve more
stable performance. In other words, similar to previous autoencoder variants~\cite{hinton2006reducing,lee2008sparse,vincent2008extracting}, our mechanism can also be used to build a deep network through model stacking. For the first layer in the deep learning model, we find the optimal layer by minimizing the objective function in Eq.~\ref{cost} using the stochastic gradient descent technique. The representations learned by the first layer are then used as the input of the second layer, and so on so forth.

\subsection{Breakpoint Detection}
After we extract the representative features of the input data through the deep learning technique described above, we calculate the distance between two features corresponding to consecutive time windows. For the $t$-th timestamp, the distance between the consecutive features ${\bm{f}}_t$ and ${\bm{f}}_{t-1}$ can be computed as
\begin{equation}\label{distance}
{Dist}_t = \frac{||{\bm{f}}_t-{\bm{f}}_{t-1}||_2}{\sqrt{||{\bm{f}}_t||_2\times||{\bm{f}}_{t-1}||_2}}
\end{equation}
where the numerator is the Euclidean distance~\cite{deza2009encyclopedia} between features corresponding to consecutive time windows, and the denominator serves as a normalization term. 

Based on the computed distance of $\{{Dist}_t\}_{t=1}^{T/N_w}$ in Eq.~\ref{distance}, we construct a distance curve and select all the peaks (local-maximal) in the curve as breakpoints detected by our approach (see details in Figure~\ref{fig:pipeline}).

We summarize our overall process for automatic breakpoint detection in Algorithm~\ref{alg1}. It is worth noting that our approach can be broadly applied for general changepoint detection even outside of its application to breakpoint detection as considered in this paper. Compared with the state-of-the-art methods~\cite{ray2002bayesian,adams2007bayesian,barry1993bayesian,bai1997estimation,erdman2008fast}, our approach 1) has no assumption on the generative models for the input data and 2) the features are automatically extracted from the input data making them representative for analysis. Through extensive experimental analysis in Section~\ref{experiment}, we will show that our method significantly outperforms previous approaches.

\section{Evaluation}\label{experiment}
In this section, we aim to validate the effectiveness of our deep learning based breakpoint detection 
method. We apply our method to real-world data sets including the Crowdsignal.io sensor data set~\cite{crowdsignals}, an EEG eye state data set~\cite{UCIEEG}, the
UCI human activity recognition data set~\cite{anguita2013public},
and the DCASE2016 sound data set~\cite{dcase_data}. We
systematically analyze the robustness of our approach under different parameter settings, with respect to the \emph{window size}, \emph{codebook (feature set) size}, and \emph{the number of layers in autoencoder models}, in order to provide practical hyperparameter selection guidelines.
Furthermore, we show the significant advantage our method has over the state-of-the-art techniques.

\begin{figure*}[!t] \centering
\subfigure[]{
\label{fig:ydata_window} 
\includegraphics[width=2.3in,height=1.5in]{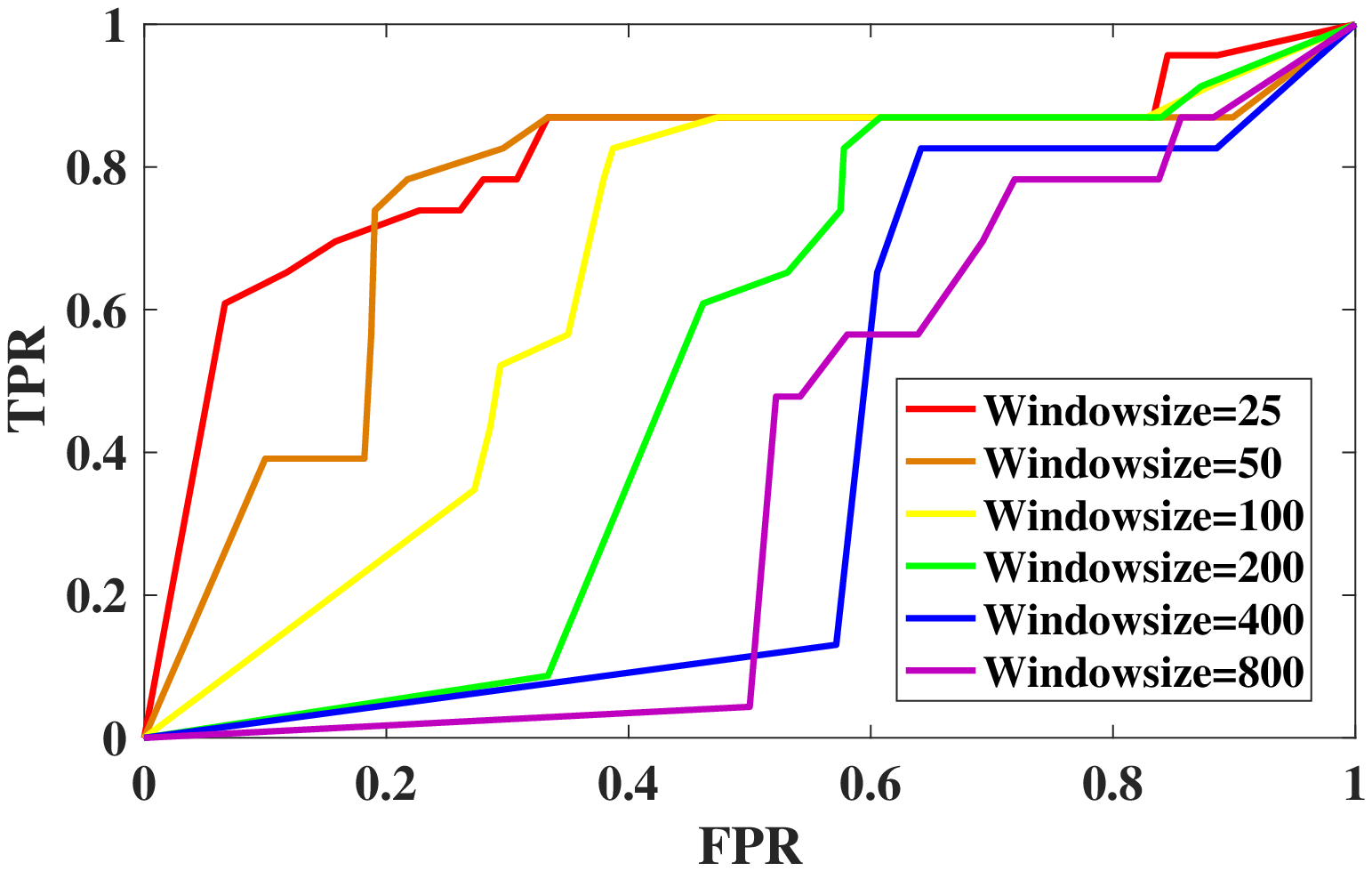}}
\hspace{-0.2in}
\subfigure[]{
\label{fig:uci_window} 
\includegraphics[width=2.3in,height=1.5in]{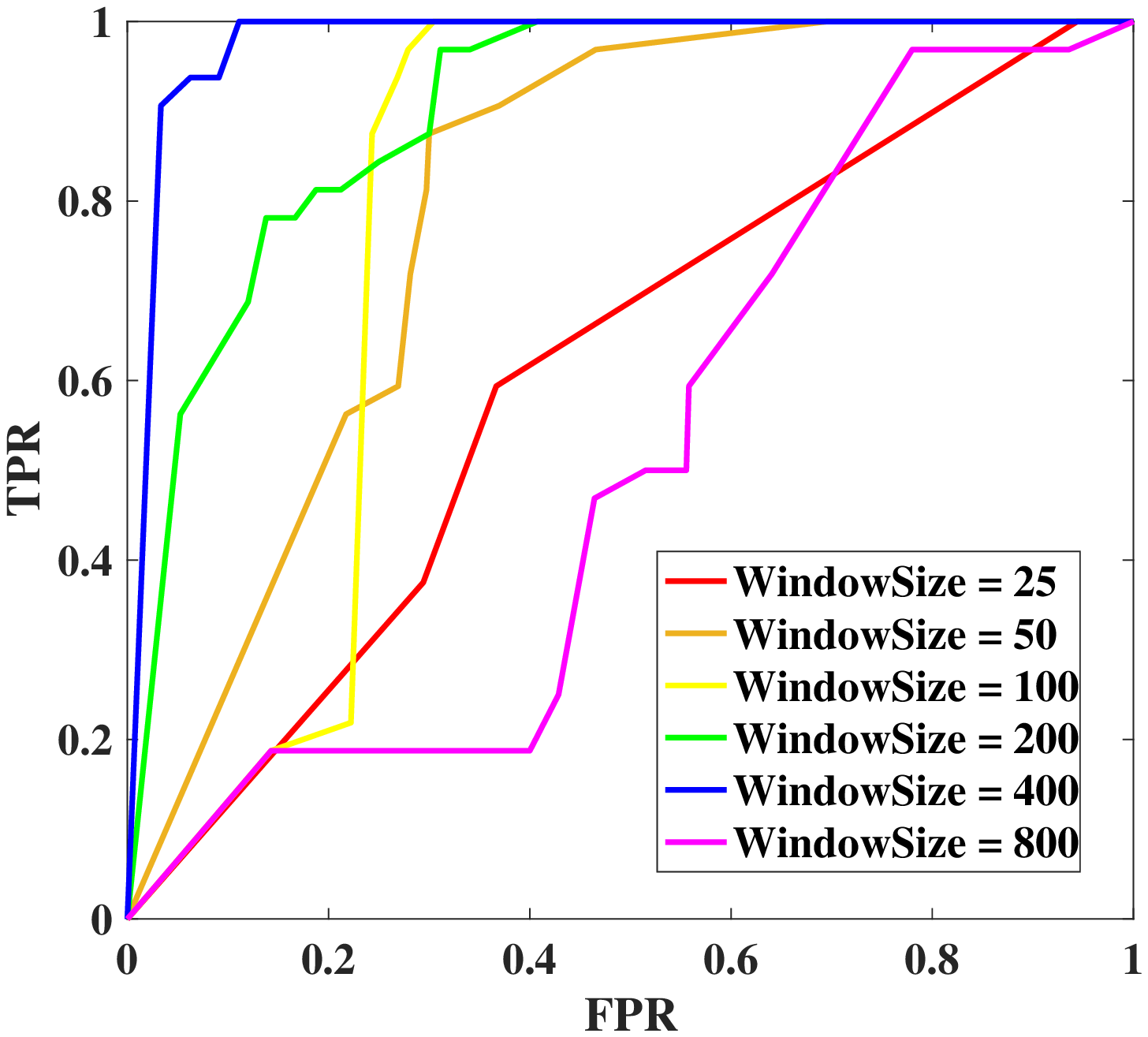}}
\hspace{-0.2in}
\subfigure[]{
\label{fig:dcase_window} 
\includegraphics[width=2.3in,height=1.5in]{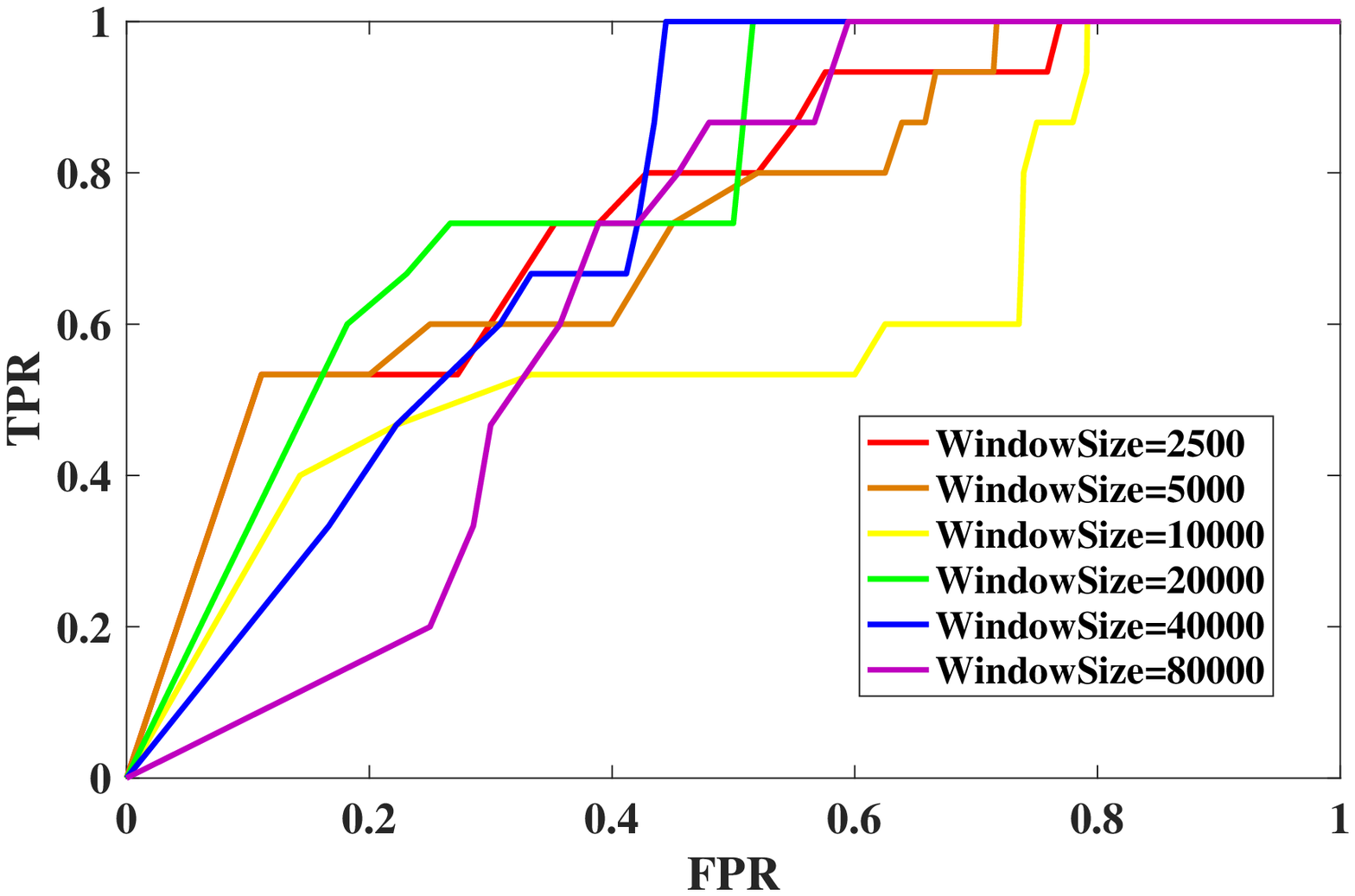}}
\caption{The ROC curve of EEG data set, UCI data set and DCASE data set under different window size $N_w$. The best performance occurs at $N_w=25$ in EEG data set, $N_w=400$ in UCI data set, and $N_w=20000$ in DCASE data set, respectively.}
\label{fig:window}
\end{figure*}

\begin{figure*}[!t] \centering
\subfigure[]{
\label{fig:ydata_statistics} 
\includegraphics[width=2.3in,height=1.5in]{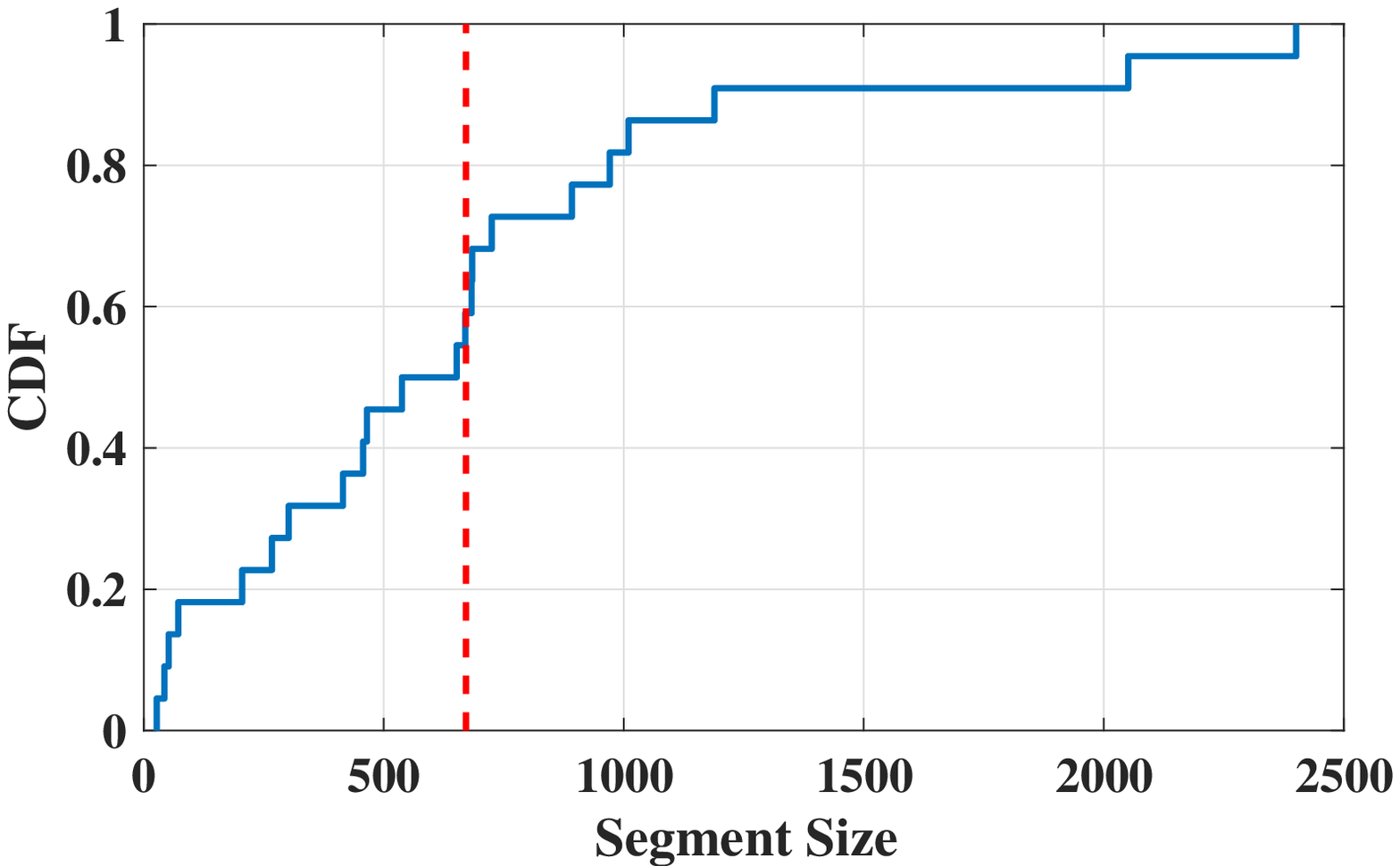}}
\hspace{-0.2in}
\subfigure[]{
\label{fig:uci_statistics} 
\includegraphics[width=2.3in,height=1.5in]{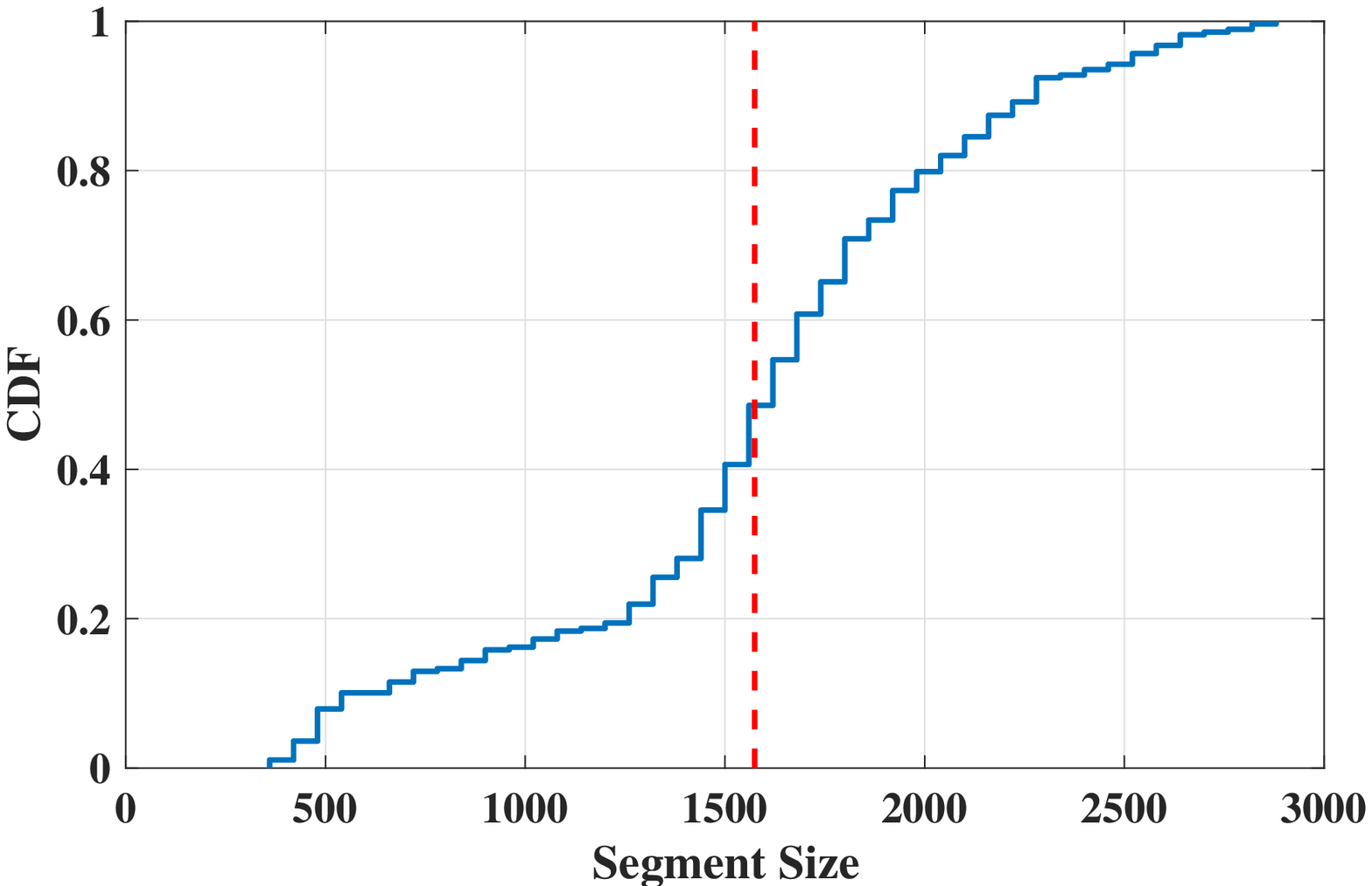}}
\hspace{-0.2in}
\subfigure[]{
\label{fig:dcase_statistics} 
\includegraphics[width=2.3in,height=1.5in]{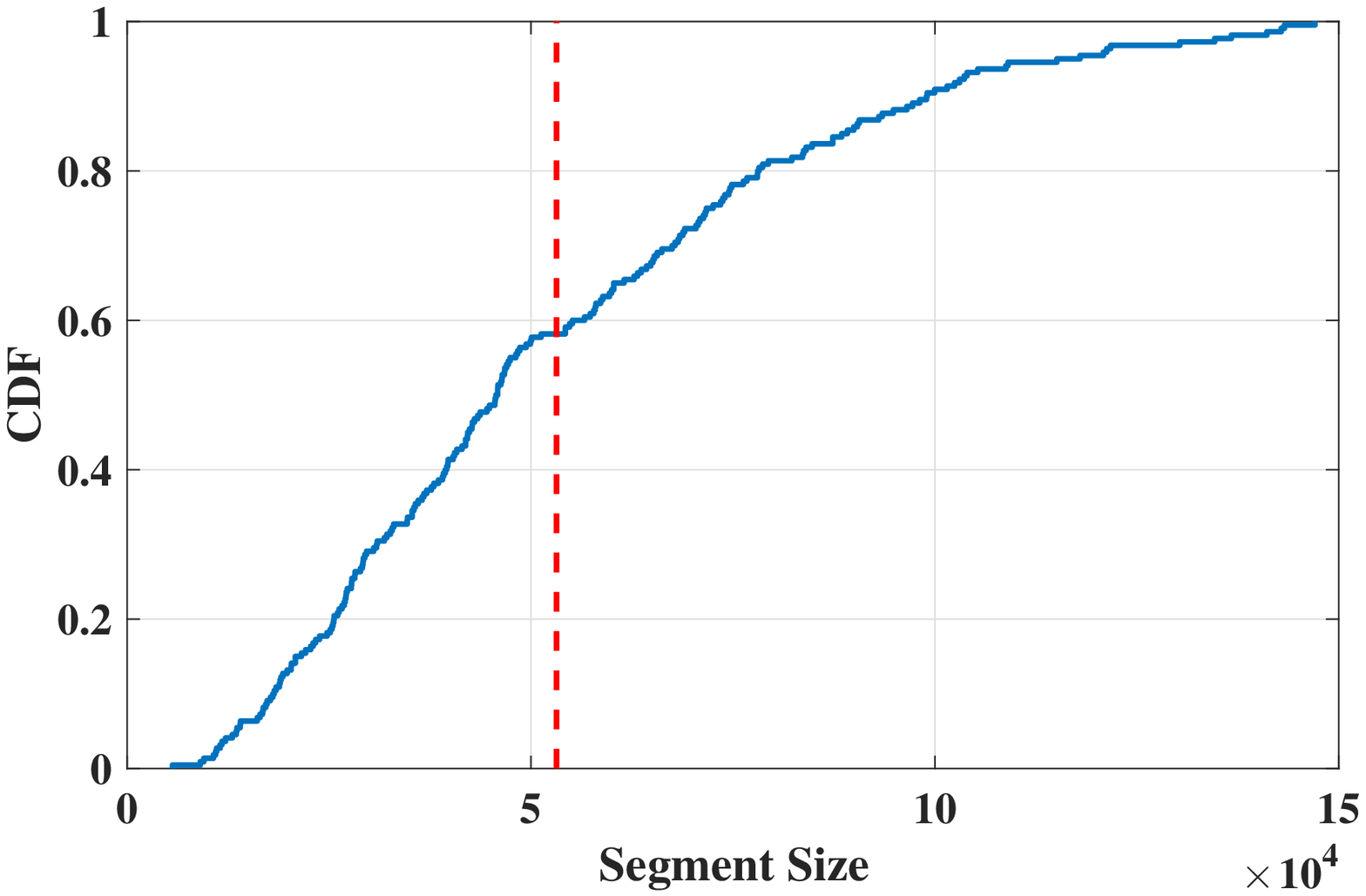}}
\caption{The CDF of the true segment sizes of the EEG data set, UCI data set and DCASE data set, respectively. The red dot line represents the average size of true segments for each data.}
\label{fig:statistics}
\end{figure*}

\subsection{Fundamental Intuition}
We demonstrate the fundamental intuition of our method for detecting both statistically-detectable changepoints (using a synthetic data) and human-specified breakpoints (using the real-world Crowdsignal.io sensor data set \cite{crowdsignals}), as shown in Figure \ref{fig:data3}. For the synthetic data, we generate a number of changepoints from a uniform distribution and each segment is generated by sampling from an exponential distribution whose parameter is sampled from a uniform distribution. The Crowdsignal.io (CI) sensor data set \cite{crowdsignals} contains mobile sensor recordings associated with users' activity information -- taking an elevator, riding an escalator and walking. 

In Figure \ref{fig:data3}, the upper figures describe the raw input signals (shown as green lines) where the red lines represent the ground truth of changepoints. The bottom figures show the distance between features in two consecutive windows (shown as blue lines and recall $Dist_t$ in Eq.~\ref{distance}), where the red dashed lines represent our detected changepoints. From Figure~\ref{fig:data3}, we show that higher distance measurements in our method correspond to ground-truth changepoints in the raw data, thus resulting in accurate changepoint detection, which lays the foundations of our approach for real-world applications.

\subsection{Evaluation on Real Traces}
We further demonstrate the effectiveness of our method by using three more real-world data sets (EEG eye state data set~\cite{UCIEEG}, UCI human activity recognition data set~\cite{anguita2013public} and the DCASE2016 sound data set \cite{dcase_data}).

The EEG eye state data set~\cite{UCIEEG} is constructed from one continuous EEG measurement with the Emotiv EEG Neuroheadset. The duration of the measurement is 117 seconds. 
Eye state is classified using camera during the EEG measurement phase and manually added to the file after analyzing the video. 
A `1' indicates the eye-closed and `0' the eye-open state. All values are in chronological order with the first measured value at the top of the data.

The UCI human activity recognition data set~\cite{anguita2013public} contains activity mode recordings carried out with a group of 30 volunteers within an age bracket of 19-48 years. Each person performed six activities wearing a smartphone (Samsung Galaxy S2) on their waist. Using its embedded accelerometer and gyroscope, it captures 3-axial linear acceleration and 3-axial angular velocity at a constant rate of 50Hz. The experiments are video-recorded to generate labels manually. 

The DCASE2016 sound data set \cite{dcase_data} contains sounds (44.1 kHz) that carry a large amount of information about our everyday environment and physical events that take place in it. Humans identify different sounds scenes (busy street, office, etc.), and recognize individual sound sources (car passing by, footsteps, etc.). The data set contains 11 classes of sound events. Each class is represented by 20 recordings. We choose a time-series data set by randomly selecting sound events.

\begin{figure*}[!t] \centering
\subfigure[]{
\label{fig:ydata_depth} 
\includegraphics[width=2.3in,height=1.5in]{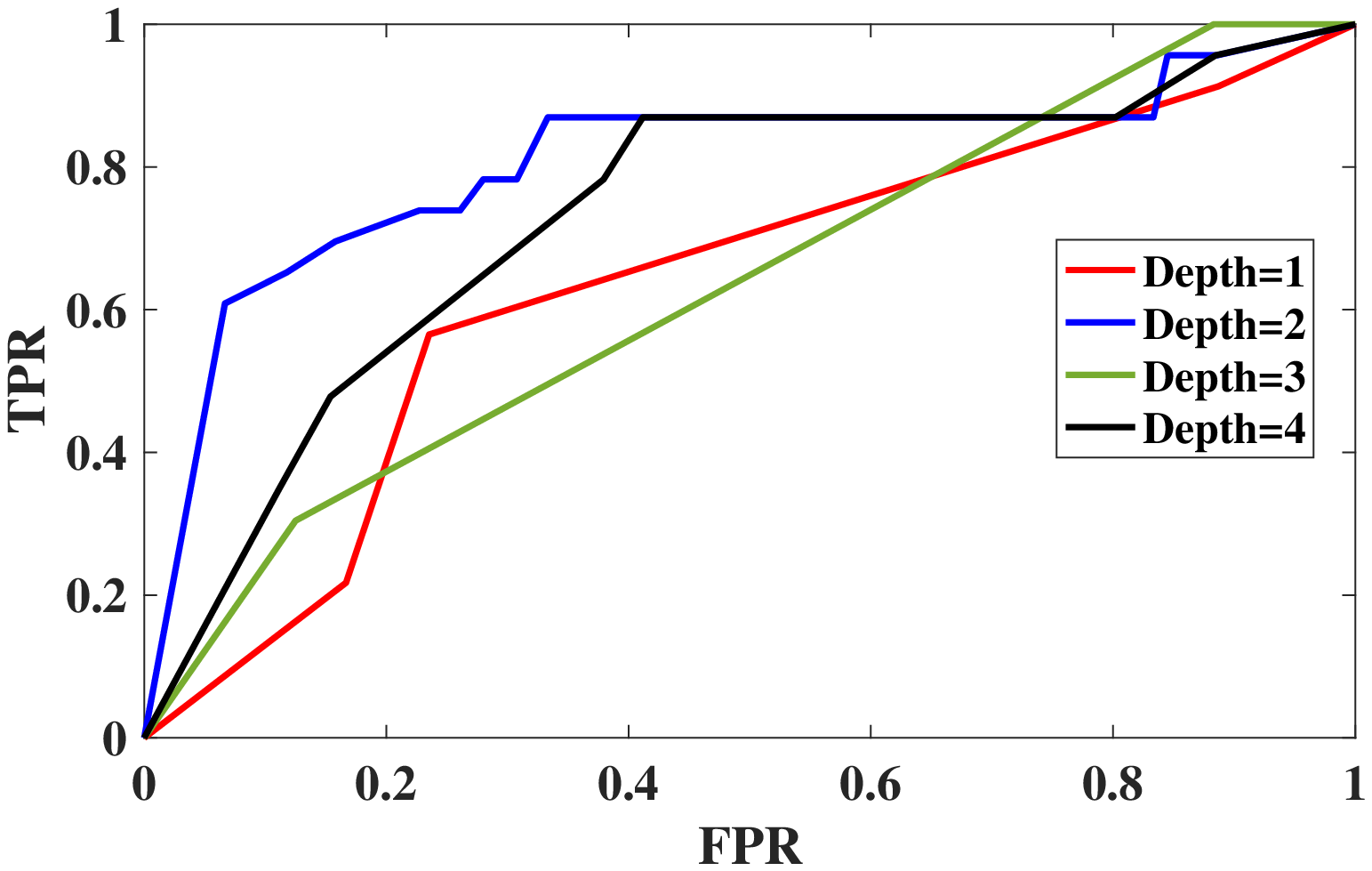}}
\hspace{-0.2in}
\subfigure[]{
\label{fig:uci_depth} 
\includegraphics[width=2.3in,height=1.5in]{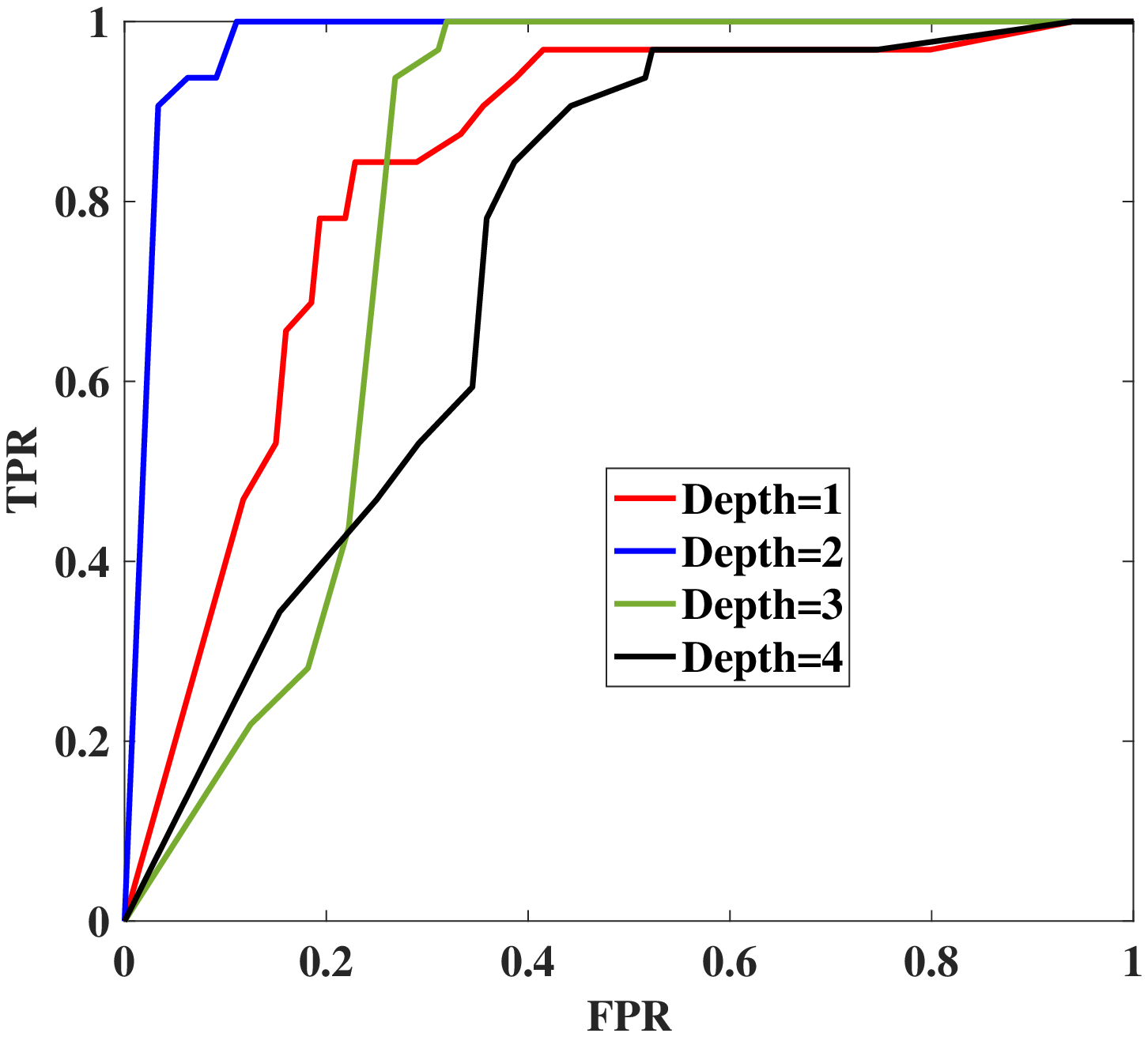}}
\hspace{-0.2in}
\subfigure[]{
\label{fig:dcase_depth} 
\includegraphics[width=2.3in,height=1.5in]{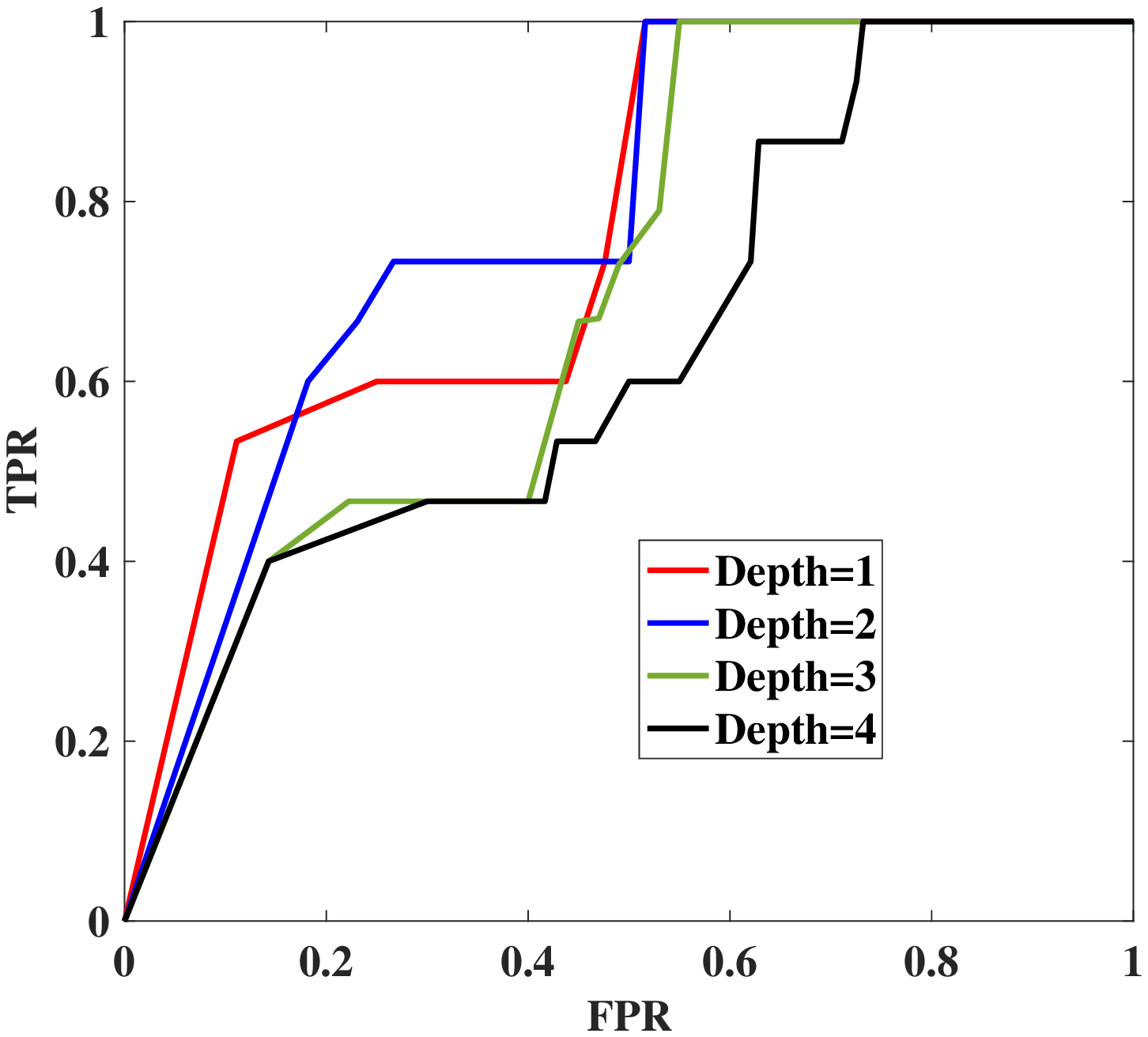}}
\caption{The ROC curve of EEG data set, UCI data set and DCASE data set under different model depth. From the experimental results, we can observe that the best detection performance is achieved with two hidden layers. Similar observations have been found in deep learning community where the two hidden layers are most commonly used~\cite{university1988continuous}.}
\label{fig:depth}
\end{figure*}

\begin{figure*}[!t] \centering
\subfigure[]{
\label{fig:ydata_code} 
\includegraphics[width=2.3in,height=1.5in]{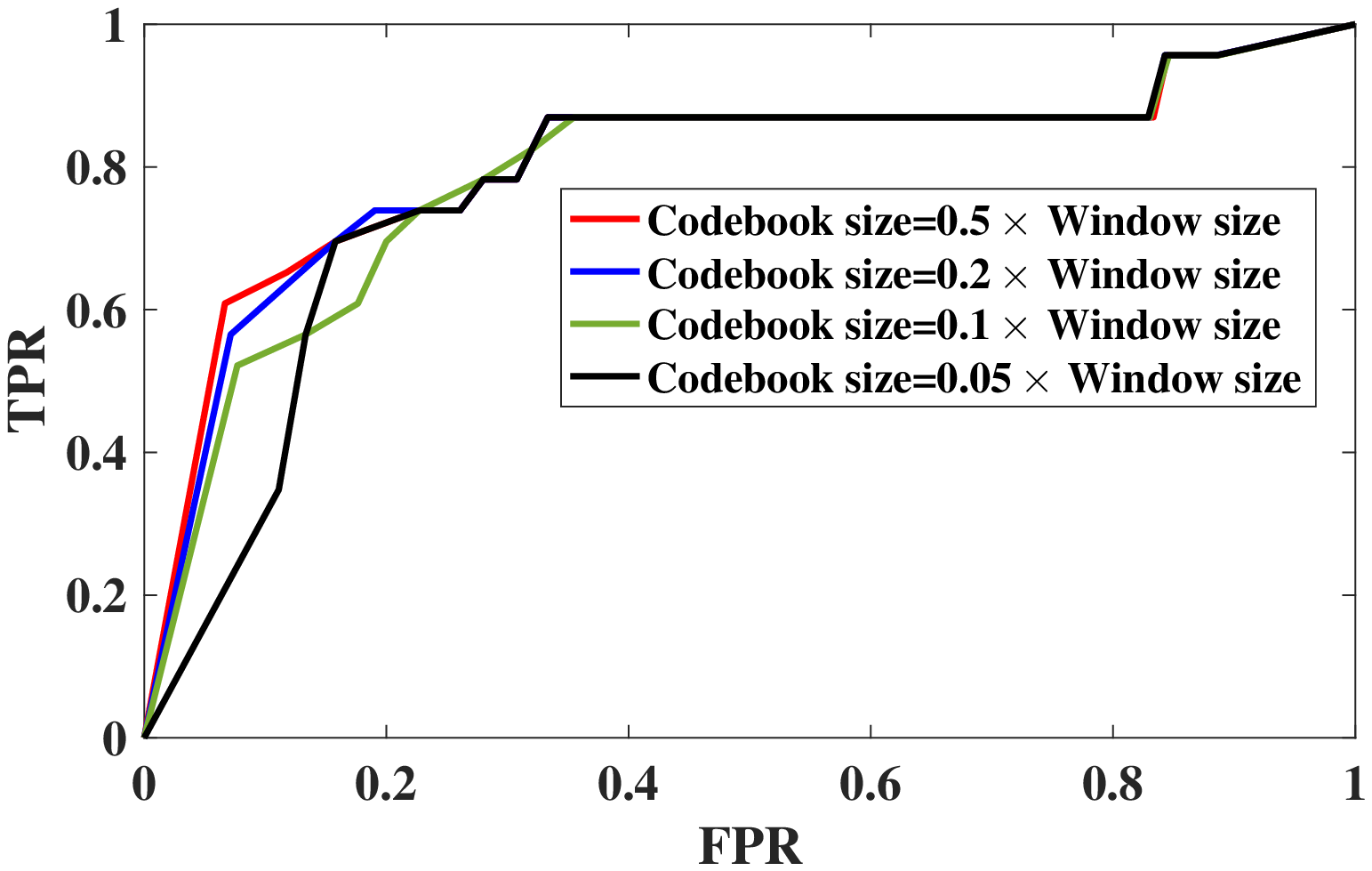}}
\hspace{-0.2in}
\subfigure[]{
\label{fig:uci_code} 
\includegraphics[width=2.3in,height=1.5in]{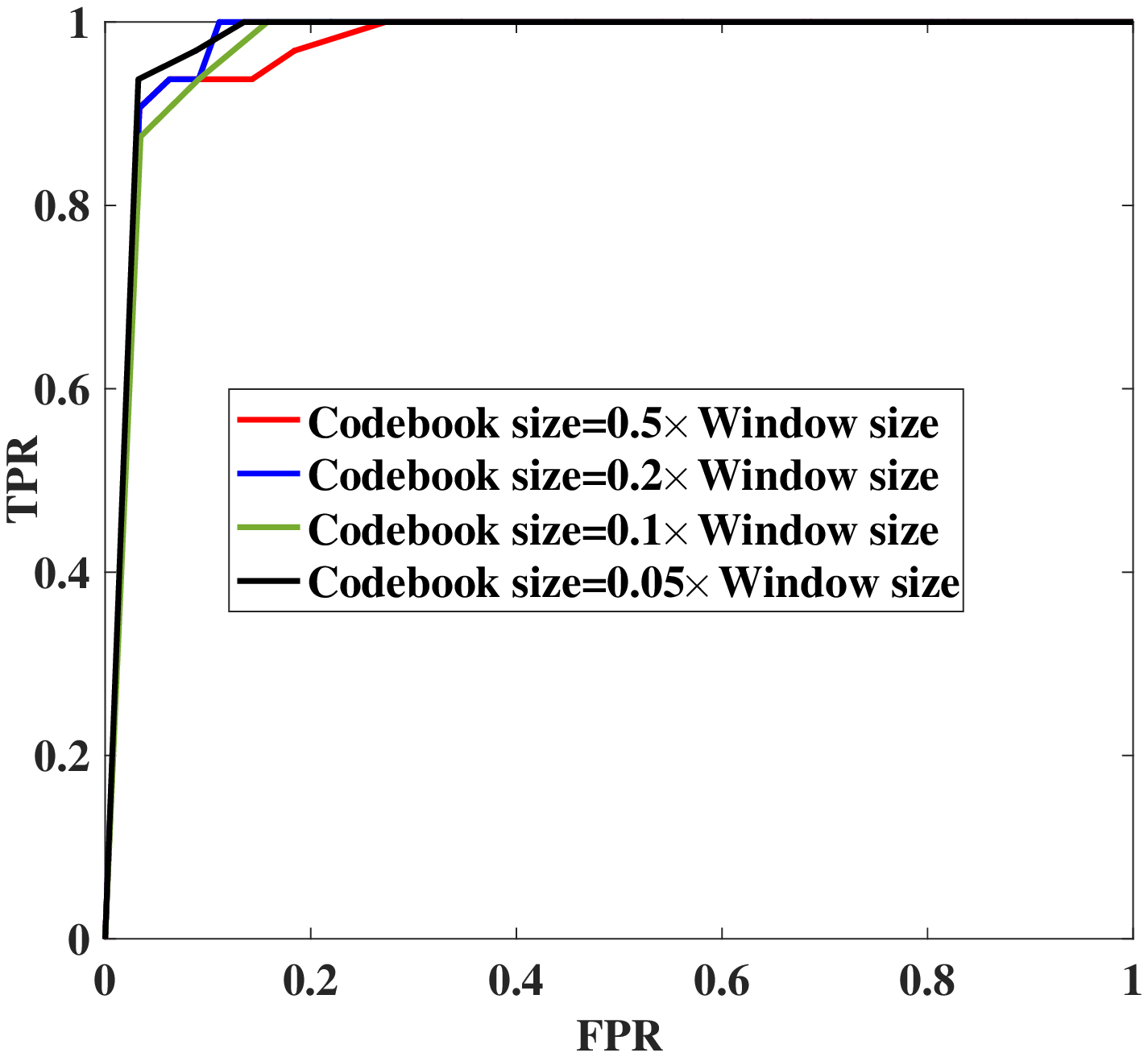}}
\hspace{-0.2in}
\subfigure[]{
\label{fig:dcase_code} 
\includegraphics[width=2.3in,height=1.5in]{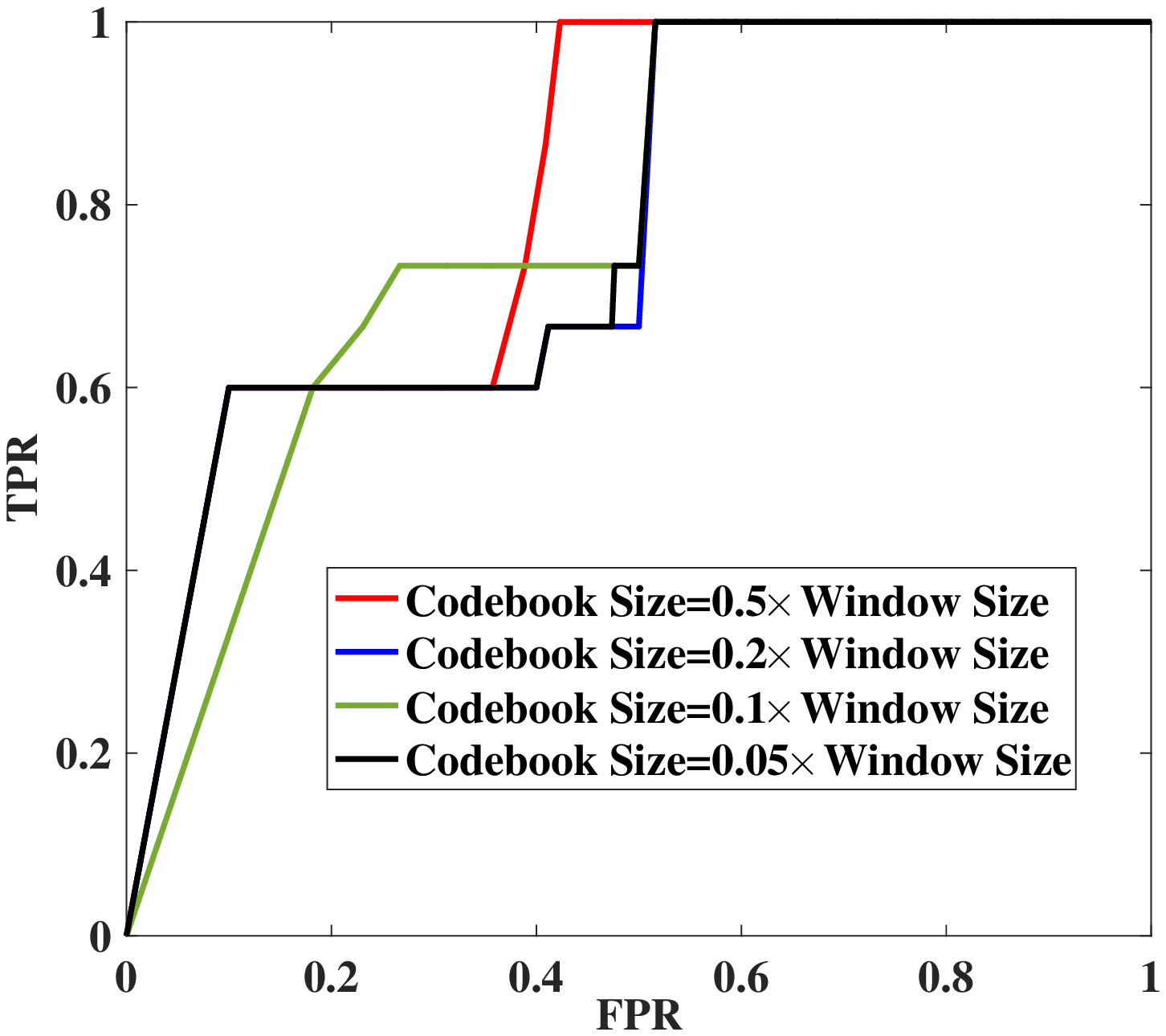}}
\caption{The ROC curve of EEG data set, UCI data set and DCASE data set under different codebook (feature set) size. We can observe that the size of the codebook has negligible influence on the performance of our detection method.}
\label{fig:code}
\end{figure*}

\subsubsection{Quantification Metrics}
\label{sec:metrics}
For fair comparsion with the state-of-the-art techniques, we use the \emph{receiver operating characteristic (ROC) curve} to measure the performance of our approach. The true positive rate and false positive rate in the ROC curve is defined as follows (similar applications can also be found in \cite{liu2013change,kawahara2012sequential}).

\begin{equation}
\begin{aligned}
\mathrm{True~Positive~Rate~(TPR)}&=\frac{N_{\bf CR}}{N_{\bf{GT}}}\\
\mathrm{False~Positive~Rate~(FPR)}&=\frac{N_{\bf AL}-N_{\bf CR}}{N_{\bf{AL}}}
\end{aligned}
\end{equation}

where $N_{\bf CR}$ denotes the number of times that the breakpoints are correctly detected, $N_{\bf GT}$ denotes the number of ground-truth breakpoints, and $N_{\bf AL}$ is the number of all detection alarms. 

Let us define a toleration distance $\tau$. For each detected breakpoint $a\in {\bf AL}$, and its corresponding closest true breakpoint $b\in {\bf{GT}}$ (i.e., $b = argmin_{i\in {\bf GT}} |a-i|$), the detected breakpoint $a$ can be seen as a \emph{correctly} detected breakpoint (i.e., $a \in N_{\bf CR}$) if the following two conditions are satisfied: {\bf Condition 1)}:  $a$ is the closest detected breakpoint of $b$ (i.e., $a = argmin_{j\in {\bf AL}} |b-j|$) and {\bf Condition 2)}: the time distance between $a,b$ is smaller than the toleration distance, i.e., $|a-b|<\tau$. We obtain the ROC curve for our method through varying the toleration distance $\tau$.

\begin{figure*}[!t] \centering
\subfigure[]{
\label{fig:ydata_compare} 
\includegraphics[width=2.3in,height=1.5in]{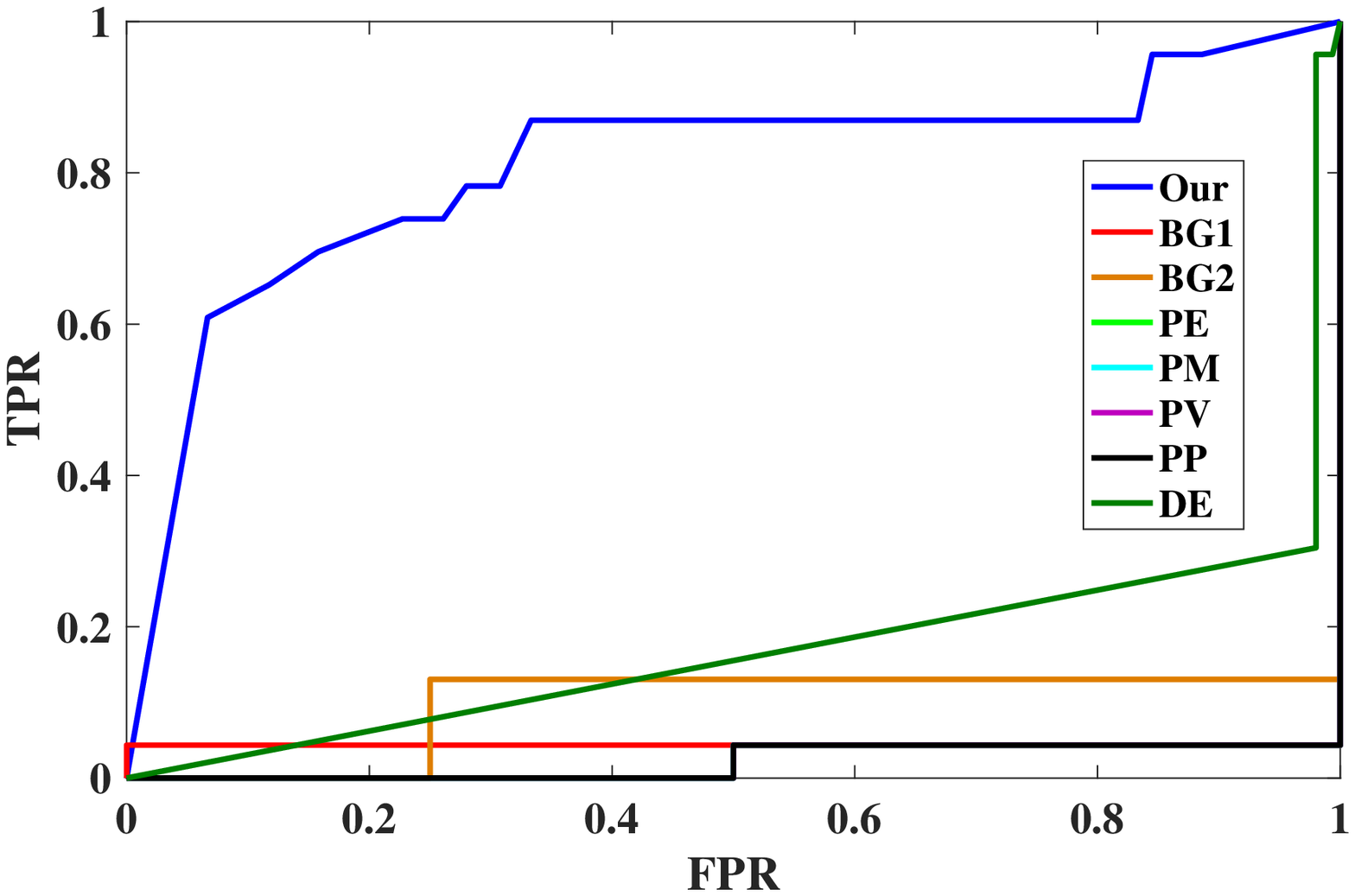}}
\hspace{-0.2in}
\subfigure[]{
\label{fig:uci_compare} 
\includegraphics[width=2.3in,height=1.5in]{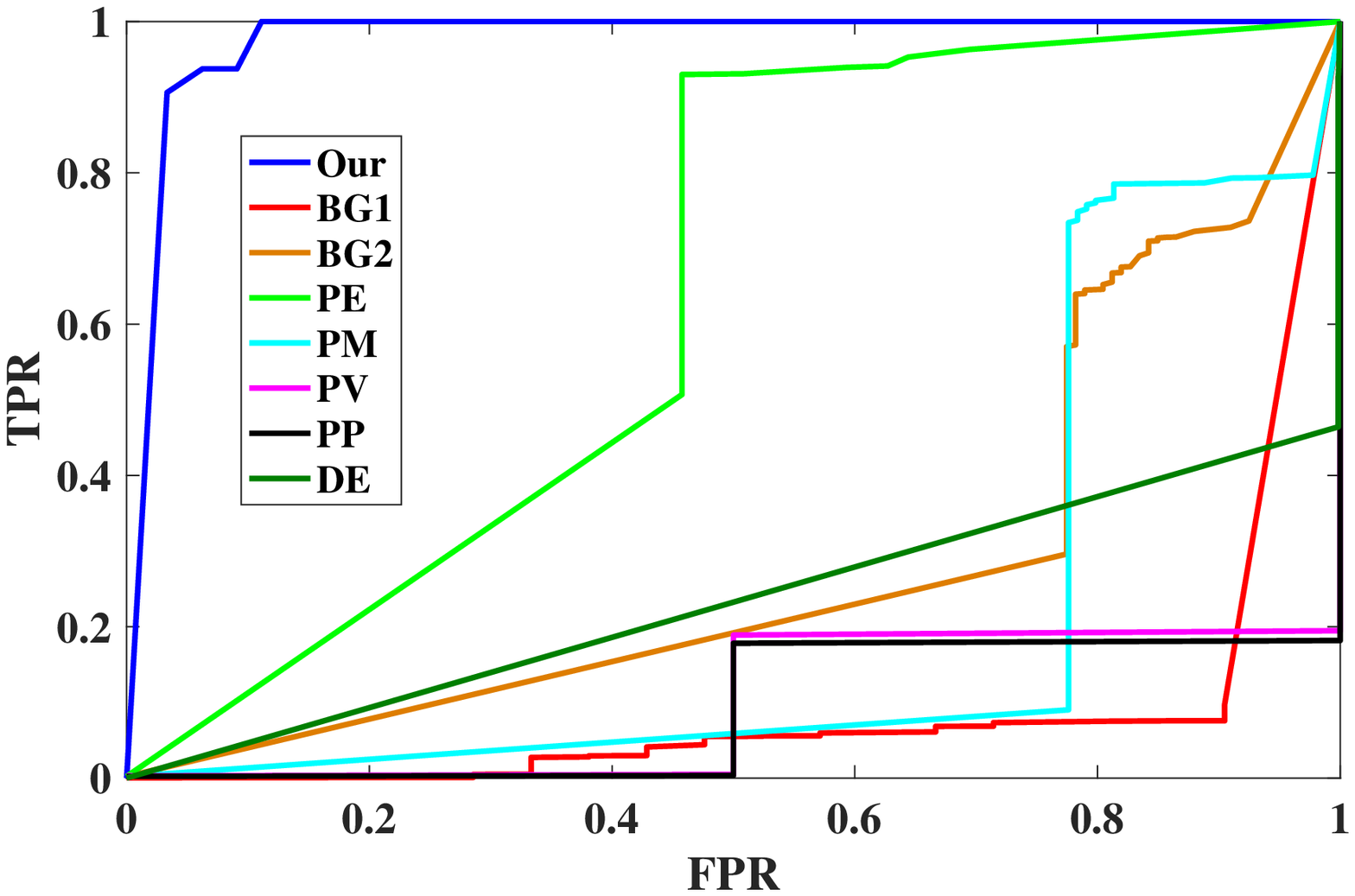}}
\hspace{-0.2in}
\subfigure[]{
\label{fig:dcase_compare} 
\includegraphics[width=2.3in,height=1.5in]{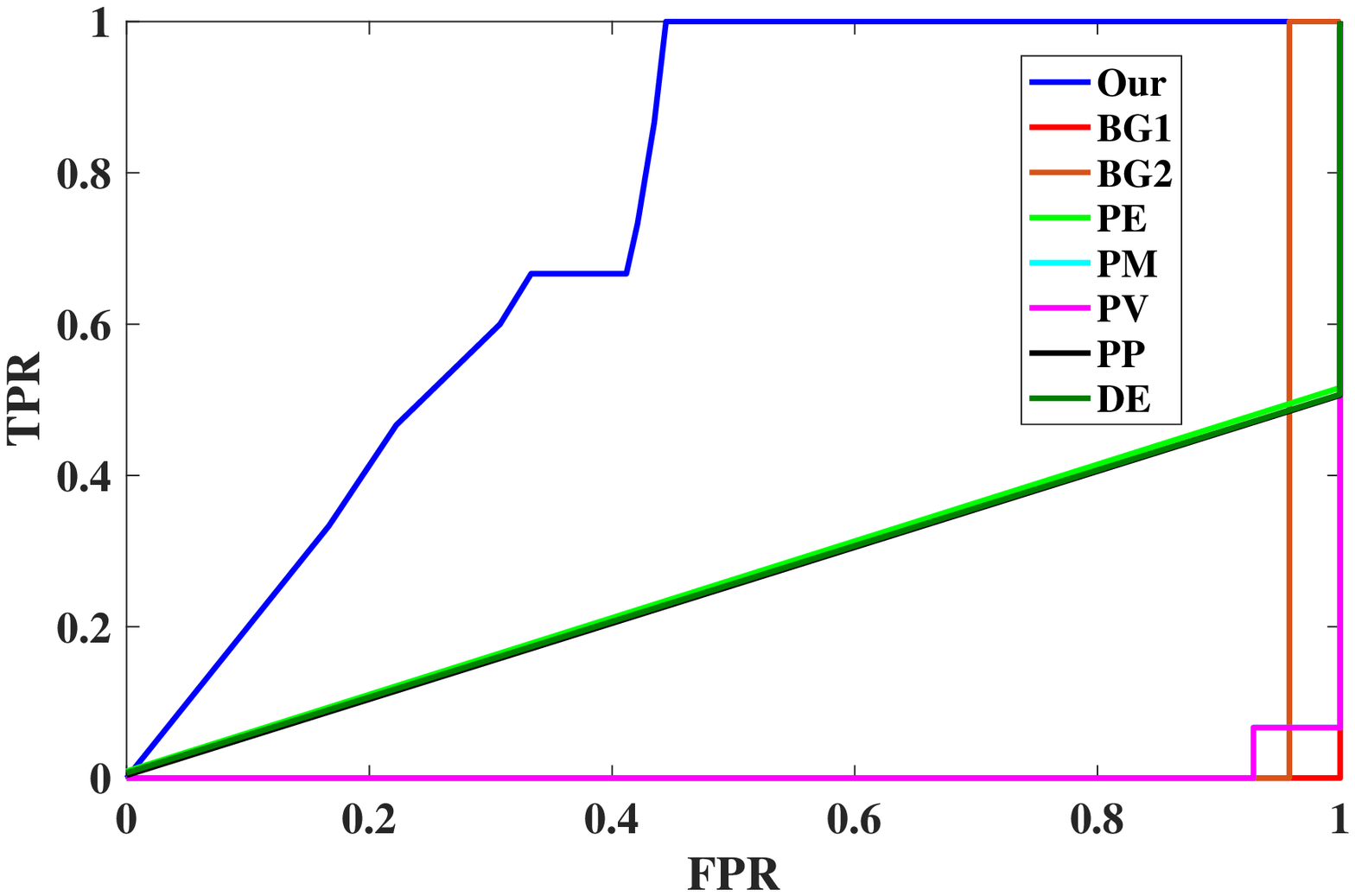}}
\caption{The ROC curve of EEG data set, UCI data set and DCASE data set under different methods. \emph{BG1 and BG2} represent online Bayesian method proposed by Adams et al.~\cite{adams2007bayesian} with prior distribution of Gamma and Gaussian, respectively. \emph{PE, PM, PV, PP} represent the Pruned Exact Linear Time (PELT) method  proposed by Killick et al.~\cite{pelt} with prior distribution of Exponential, Mean-shift, Mean-variance shift, and Poisson, respectively. 
\emph{DE} represents the density-ratio estimation method proposed by Liu et al.~\cite{liu2013change}. From the experimental results, we can observe that our approach significantly outperforms previous techniques.}
\label{fig:compare}
\end{figure*}

\subsubsection{Sensitivity Analysis} \label{sec:sensitivity}
We examine the sensitivity of our method under different parameter settings in order to provide practical guidelines for hyperparameter selection.

\noindent{\bf{Effects of Window Size:}} First, we show the sensitivity of our approach with respect to the window size $N_w$, by setting the depth of the model to two and the ratio between the codebook (feature set) size and the input data size as $\dim({\bm{f}}_t)/\dim({\bm{s}}_t)=0.1$. Figure \ref{fig:window} shows the ROC curve of each data set using different window sizes, where each curve is generated by calculating the false-positive and corresponding 
true-positive rate as we vary the guard band around the true breakpoint, 
as described in Section~\ref{sec:metrics}.

From Figure~\ref{fig:window}, we observe that the best window size (for achieving the best breakpoint detection performance) is 25 for EEG data set, 
400 for UCI data set, and 20000 for DCASE data set. To investigate how to select the best window size for a data set, we show the relationship between the true segment distribution of the input data and its corresponding best window size. In Figure~\ref{fig:statistics}, we describe the cumulative distribution function (CDF) of true segment sizes of the input data, where the red line is the average size of all the true segments. By comparing Figure~\ref{fig:window} and Figure~\ref{fig:statistics}, we observe that the best window size for a data set is roughly the true segment size corresponding to $CDF=0.1$. Therefore, our experiments suggest that the best window size should be set to the size which corresponds to $CDF=0.1$ in the true segment distribution. 

\noindent{\bf{Effects of Model Depth:}} Since we use stacked autoencoders to achieve stable detection results (recall Section~\ref{autoencoder}), we further show how its depth influences our method's performance. We set the ratio between the codebook (feature set) size and the input data size as $\dim({\bm{f}}_t)/\dim({\bm{s}}_t)=0.1$ and the window size to $25$ for EEG data set, $400$ for UCI data set, and $20000$ for DCASE data set, respectively (according to Figure~\ref{fig:window}). 
From the experimental results as shown in Figure \ref{fig:depth}, we can find that the best detection performance is achieved with two hidden layers in the autoencoder models for all the three data sets. Similar observations have been made in related
work~\cite{university1988continuous} and the two hidden layers are most commonly used in existing deep learning research.

\noindent{\bf{Effects of Codebook (Feature Set) Size:}} We further examine the effect of the codebook (feature set) size on our method. We set the model depth to two (according to Figure~\ref{fig:depth}) and the window size to $25$ for EEG data set, $400$ for UCI data set, and $20000$ for DCASE data set (according to Figure~\ref{fig:window}). Figure \ref{fig:code} shows the ROC curve under different codebook (feature set) sizes for the three data sets. We can see that the size of the codebook only has negligible influence on the detection performance of our method.

\noindent{\bf Hyperparameter Selection Heuristic:} Figures~\ref{fig:window},~\ref{fig:depth},~\ref{fig:code}, provide a guide for choosing hyperparameters in our method: 1) set the window size to the size corresponding to $CDF=0.1$
in the true segment size distribution; 2) set the model depth as $2$; and 3) randomly select the codebook (feature set) size. The performance of our method can be further improved by feeding the detection results back into the network, which will fine-tune these hyperparameters automatically. 
This technique will be explored in future work.

\subsection{Comparison with Existing Methods}\label{sec:compare}
We further compare our method with existing changepoint detection techniques. Bayesian changepoint detection 
has been well studied in the literature  and several important variants have been developed~\cite{ray2002bayesian,adams2007bayesian,barry1993bayesian,bai1997estimation,erdman2008fast}. We compare our method with these approaches on each of the 
three real-world data sets, as shown in Figure~\ref{fig:compare}. \emph{BG1 and BG2} represent online Bayesian method proposed
by Adams et al.~\cite{adams2007bayesian} with prior distribution of Gamma, and Gaussian, respectively. \emph{PE, PM, PV, PP} represent the Pruned Exact Linear Time (PELT) method proposed by Killick et al.~\cite{pelt}, with prior distribution of Exponential, Mean-shift, Mean-variance shift, and 
Poisson, respectively. 
\emph{DE} represents the density-ratio estimation method proposed by Liu et al.~\cite{liu2013change}.

\begin{table*}[t!]
\centering 
\newcommand{\tabincell}[2]{\begin{tabular}{@{}#1@{}}#2\end{tabular}}
\centering
	\begin{tabular}{|c|c|c|c|c|c|c|c|c|c|c|c |}\hline
	& \multicolumn{3}{c|}{\bf EEG Data Set}  & \multicolumn{3}{c|}{\bf UCI Data Set} & \multicolumn{3}{c|}{\bf DCASE Data Set} \\ \hline
& \tabincell{c}{\bf PR} & {\bf MSE} & \tabincell{c}{\bf PL} & \tabincell{c}{\bf PR} & {\bf MSE} & \tabincell{c}{\bf PL} & \tabincell{c}{\bf PR} & {\bf MSE} &  \tabincell{c}{\bf PL} \\ \thickhline
{\bf Our Method} &0.96&5.11&\textbf{0.22} &1.06&20.04&\textbf{1.25} &1.13&12.82&\textbf{1.71} \\ \hline
{\bf BG1} &500&6.38&3183 &0.66&34.48&11.85 &0&inf&undef \\ \hline
{\bf BG2} &0.173&5.95&4.92 &4.16&16.19&51.09 &23.87&20.03&457.98  \\ \hline
{\bf PE}&0.09&19.27&17.60 &1.84&3.01&2.54 &3027.6&0.08&234.56  \\ \hline
{\bf PM}&0.09&19.27&17.60 &4.19&8.69&27.68 &0.93&133.27&8.88 \\ \hline
{\bf PV}&0.09&19.27&17.60 &0.06&29.38&27.54 &0.93&133.27&8.88  \\ \hline
{\bf PP}&0.087&19.27&17.60 &0.06&29.38&27.55 &3027.67&0.08&234.57  \\ \hline
{\bf DE}&47.96&5.78&271.8 &621&6.52&4043 &2052&0.28&576.5  \\ \hline
    \end{tabular}
    \vspace{1em}
    \caption{Comparison of existing approaches with our deep learning based method. \emph{BG1, BG2} represent online Bayesian method proposed by Adams et al.~\cite{adams2007bayesian} with prior distribution of Gamma, Gaussian, respectively. \emph{PE, PM, PV, PP} represent the Pruned Exact Linear Time (PELT) method  proposed by Killick et al.~\cite{pelt} with prior distribution of Exponential, Mean-shift, Mean-variance shift, and Poisson, respectively. 
\emph{DE} represents the density-ratio estimation method  proposed by Liu et al.~\cite{liu2013change}. We calculate the \emph{prediction ratio (PR)}, \emph{mean-squared error (MSE)}, and \emph{prediction loss (PL)} for the EEG data set, UCI data set, and DCASE data set, respectively. We can observe the significant advantage our approach has over previous methods.}
    \label{tab:comp}
\end{table*}

\subsubsection{Parameter Settings}\label{exp_parameter}
Based on the analysis in Section~\ref{sec:sensitivity}, we set the model depth to 2, the ratio between the codebook (feature set) size and the input data size as $\dim({\bm{f}}_t)/\dim({\bm{s}}_t)=0.1$, the window size to 25 for EEG data set, 400 for UCI data set, and 20000 for DCASE data set.
For fair comparison with the state-of-the-art approaches, we use the same parameter settings used in their papers \cite{adams2007bayesian, pelt, liu2013change}. Specifically, for \emph{BG1}, we use a Gamma prior on the inverse variance, with $a = 1$ and $b = 1$. The rate of the exponential prior on the segment size is 1000. For \emph{BG2}, we use a univariate Gaussian model with prior parameters $\mu = 1.15\times 10^5$ and $\sigma = 1\times 10^4$. The rate of the exponential prior is 250. For \emph{DE}, we use $n=50$, $k=10$, and $\alpha=0.1$.

\subsubsection{Experimental Results}
In Figure~\ref{fig:compare}, we observe that our method consistently achieves higher TPR's over other approaches under the same level of FPR. For instance, when the FPR equals to $0.2$, we achieve significantly higher TPR over other methods with up to $7\times$, $3\times$, and $2\times$ improvement, corresponding to the EEG data set, UCI data set and DCASE data set respectively. Therefore, our method significantly outperforms the state-of-the-art approaches. 

Table~\ref{tab:comp} compares them with a different evaluation criteria. Rather than varying the guard band to calculate a true and false positive rate (recall Section~\ref{sec:metrics}), we use the closest predicted breakpoint to each true breakpoint to calculate the \emph{mean-squared error (MSE)}.
To measure the rate of over/under prediction of breakpoints, we also calculate the \emph{prediction ratio} (PR), between the number of detected breakpoints $N_{\bm AL}$ and the number of true breakpoints $N_{\bm GT}$.
\begin{equation}
prediction~ratio = \frac{N_{\bm{AL}}}{{N_{\bm{GT}}}}
\end{equation} 

A higher prediction ratio means that the algorithm tends to over-predict while a lower one means it under-predicts.  A ratio of 1 means the algorithm predicted the exact number of actual breakpoints.  Algorithms with high prediction ratios will tend to have lower MSE's, since there is a higher probability that a predicted breakpoint will be close to an actual one. Conversely, algorithms with low prediction ratios will tend to have higher MSE's. To capture this tradeoff we introduce a new measure \emph{prediction loss} (PL) defined as follows.
\begin{equation} \label{eq:efficiency}
\begin{aligned}
prediction~loss &= \left| 1-\frac{{N_{\bm{AL}}}}{{N_{\bm{GT}}}}\right| \times MSE \\
				&= \left| 1- prediction~ratio \right| \times MSE 
\end{aligned}
\end{equation}

The smaller the \emph{prediction loss} the better the algorithm is performing. However, there are two situations where the \emph{prediction loss} does not capture the performance well: 1) when the number of predicted breakpoints is 0; and 2) when the algorithm predicts a breakpoint at each timestamp. Both
will result in perfect \emph{prediction loss} of 0. To prevent this the \emph{prediction loss} is denoted as \emph{undefined} when the number of predicted breakpoints is zero (${N_{\bm{AL}}}=0$) or we note a high \emph{prediction ratio}. From Table~\ref{tab:comp}, we observe that our deep learning based method, achieves the lowest \emph{prediction loss} among all algorithms, often producing \emph{prediction ratio} near 1 and small MSE. These experimental results further demonstrate the advantage of our approach over existing methods.

\begin{figure*}[!t] \centering
\subfigure[]{
\label{fig:ydata_pdf} 
\includegraphics[width=2.3in,height=1.5in]{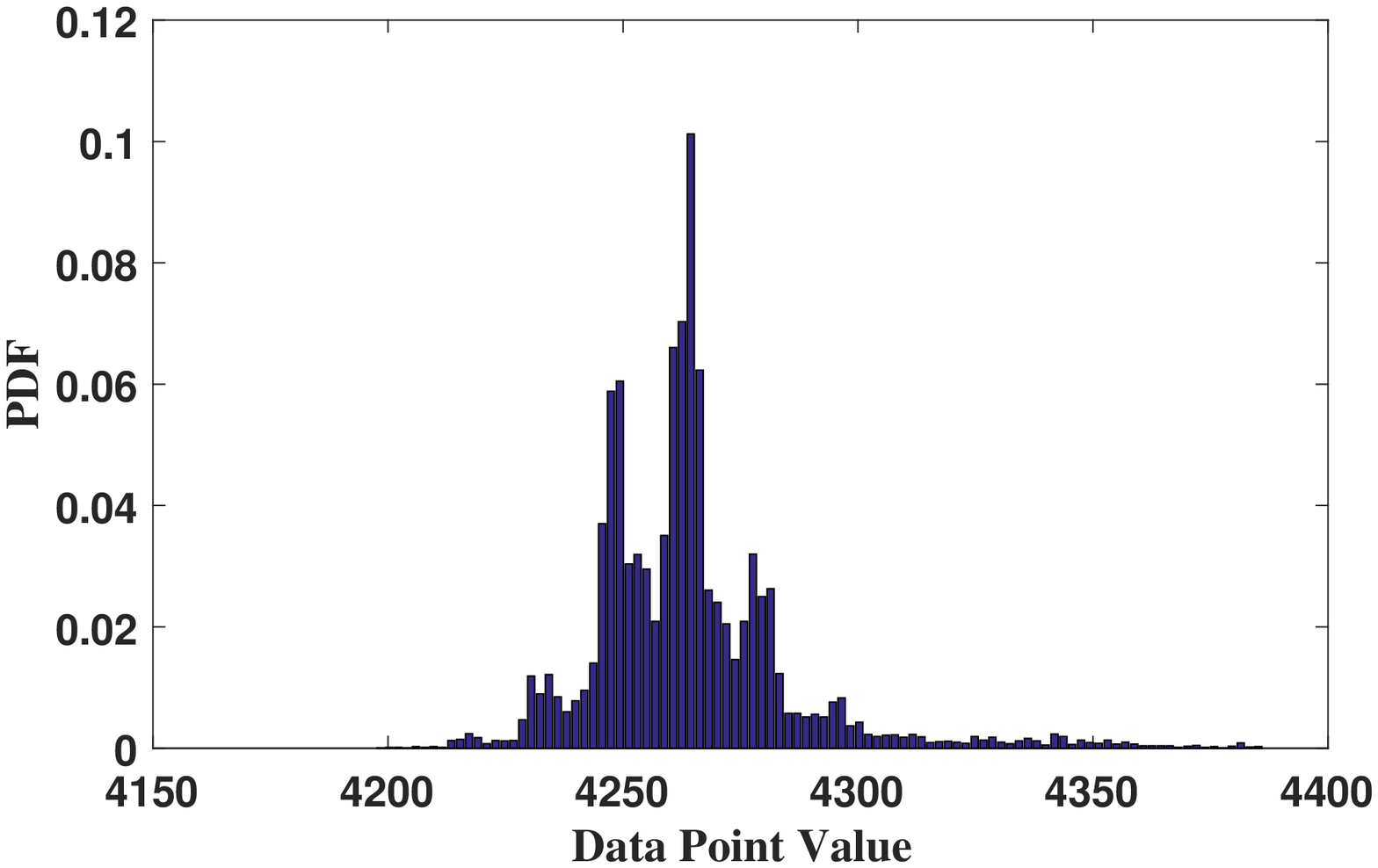}}
\hspace{-0.2in}
\subfigure[]{
\label{fig:uci_pdf} 
\includegraphics[width=2.3in,height=1.5in]{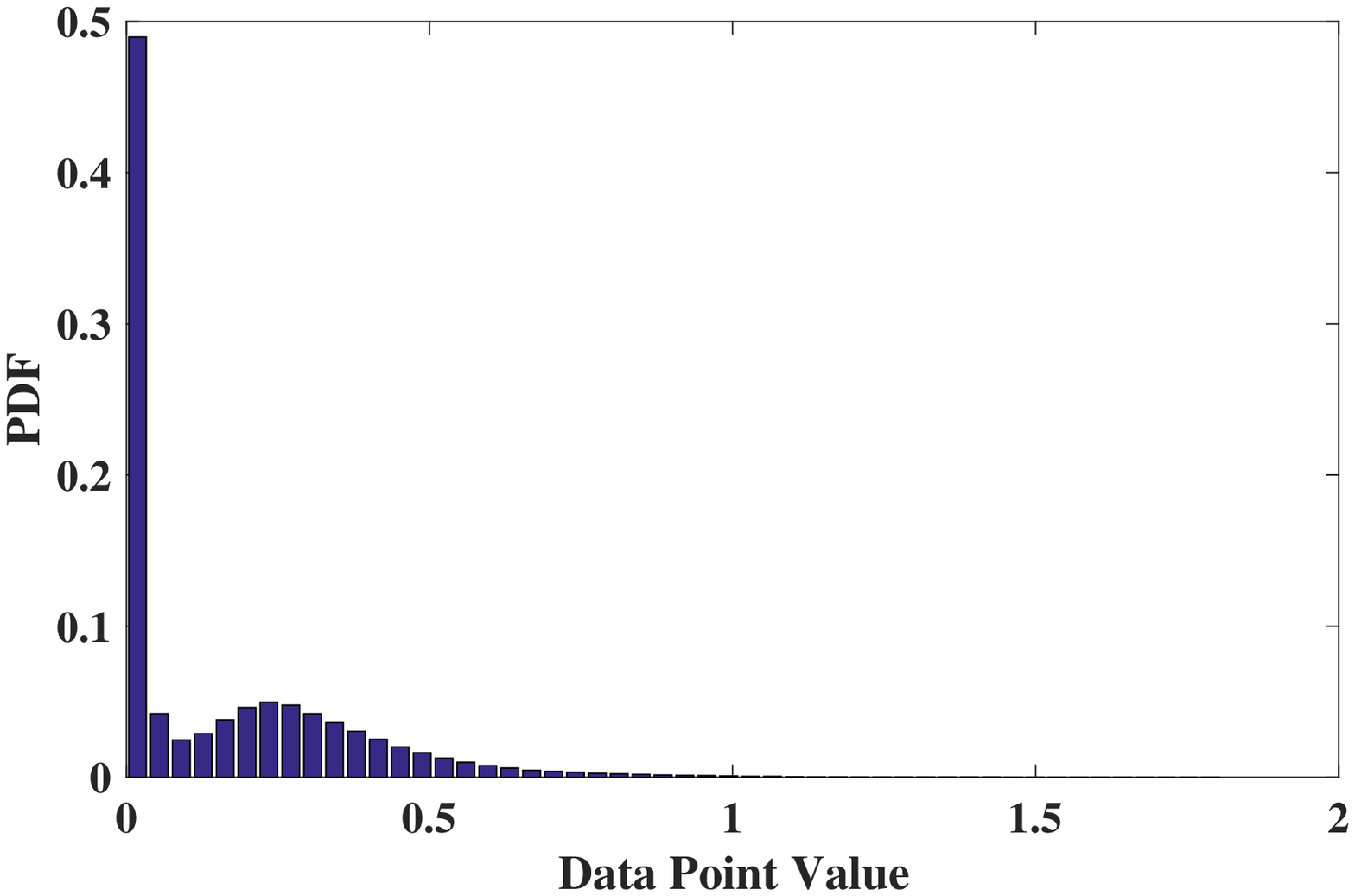}}
\hspace{-0.2in}
\subfigure[]{
\label{fig:dcase_pdf} 
\includegraphics[width=2.3in,height=1.5in]{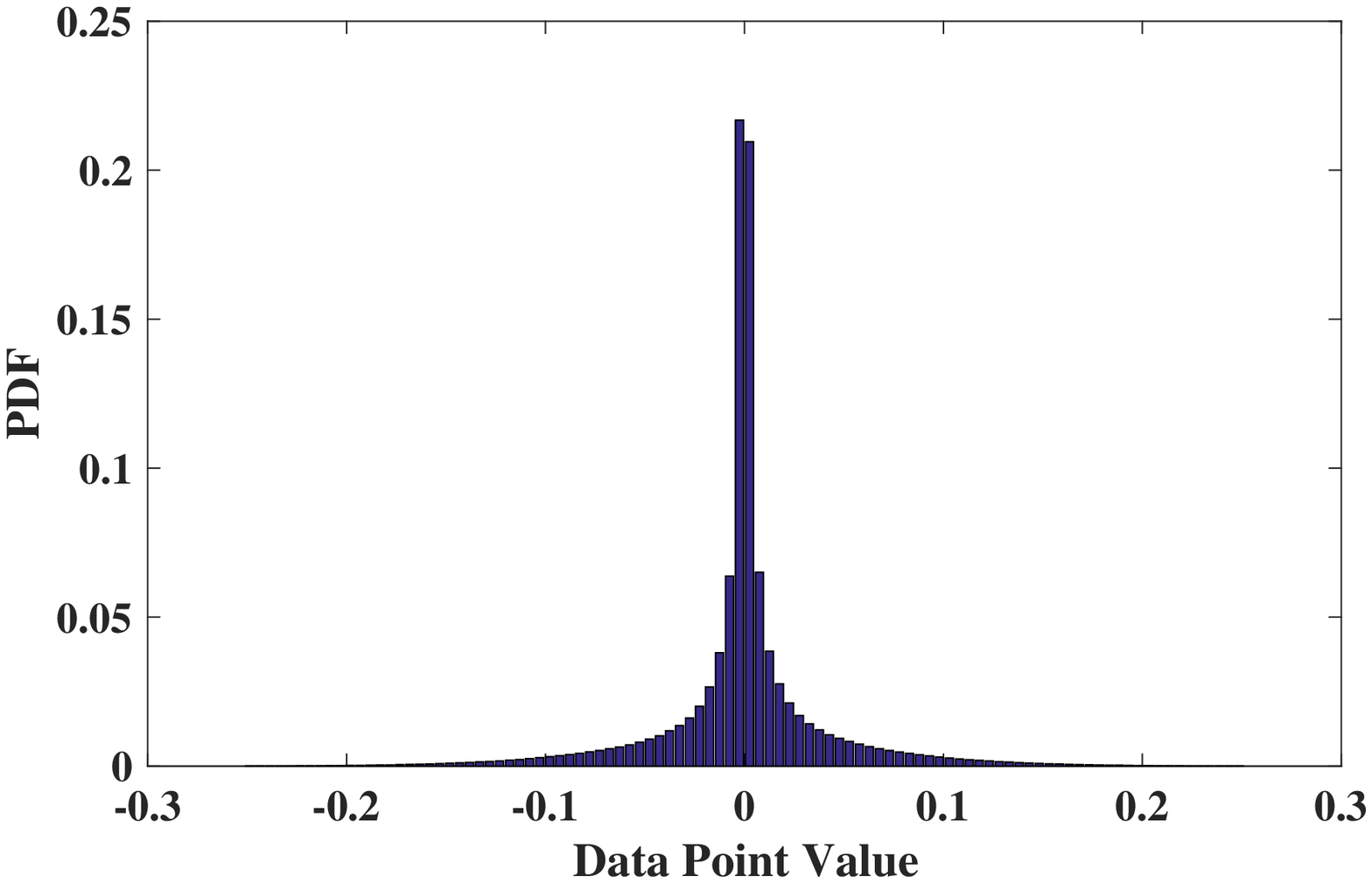}}
\caption{The PDF of the data points of the EEG data set, UCI data set and DCASE data set, respectively.}
\label{fig:pdf}
\end{figure*}

\subsection{Discussion}From Section \ref{sec:compare}, we observe that our method achieves significant improvement over the state-of-the-art approaches \cite{adams2007bayesian, pelt, liu2013change}. The reason is that these previous methods may suffer from one or several of the following limitations:
\begin{enumerate}[$\bullet$]
\item Assume that the input data points are generated from certain distributions and that each data point is independently and identically distributed (i.i.d.).
\item Assume that the segments of the input data are generated from certain distributions.
\item Lack of explicit guide for tuning the hyperparameters under different types of data sets.
\end{enumerate}

However, these assumptions may not be applicable for real-world data sets for the following reasons 1) the input data points and the segments may not be generated from certain distributions (as shown in Figure~\ref{fig:statistics} which plots the CDF of the true segment size in the three data sets, and Figure~\ref{fig:pdf} which plots the probability density function (PDF) of the data points in the three data sets); 2) the data points may be correlated with each other (e.g., smartphone sensor data corresponding to human activity modes); and 3) fixed parameter settings may not be optimal for all the data sets. For instance, the average size of true segments in the three data sets (EEG data set, UCI data set and the DCASE data set) is $671$,~$1574$ and $53156$, respectively, as shown in Figure~\ref{fig:statistics}. In order to achieve good detection performance, an optimal window size should be selected to adapt to each data set. As shown in Figure~\ref{fig:window}, the optimal window size automatically tuned in our algorithm is $25,~ 400,~ 20000$ for the three data sets respectively. 

In summary, our deep learning based method can overcome all the above limitations of previous work, by enabling the system to automatically learn the hidden structures and extract the most useful features of the time-series data. Therefore, our approach can be generally applied to a broad coverage of real-world applications.

\section{Conclusion}
In this paper, we propose a novel method to detect human-specified breakpoints through exploiting deep learning techniques for automatically extracting features that can well represent the input time-series data. It is worth noting that our approach is of independent interest for general changepoint detection even outside the context of breakpoint detection as considered in this paper. Unlike previous methods, our approach does not rely on specifying a prior generative model of the input data. Furthermore, we introduce a simple hyperparameter tuning criteria through careful sensitivity analysis on the window size, codebook size, and depth of the network. Through extensive experiments on multiple types of real-world data sets including human-activity sensing, speech, and EEG traces, we demonstrate the effectiveness of our proposed method and show that it significantly outperforms existing approaches. Our technique can serve as a key primitive for analyzing a broad range of real-world time series data.

\bibliographystyle{ACM-Reference-Format}
{\Large{
\bibliography{sigproc} }}

\end{document}